\PassOptionsToPackage{dvipsnames}{xcolor} 
\documentclass[sigconf,nonacm]{aamas}

\makeatletter
\gdef\@copyrightpermission{
  \begin{minipage}{0.2\columnwidth}
   \href{https://creativecommons.org/licenses/by/4.0/}{\includegraphics[width=0.90\textwidth]{by}}
  \end{minipage}\hfill
  \begin{minipage}{0.8\columnwidth}
   \href{https://creativecommons.org/licenses/by/4.0/}{This work is licensed under a Creative Commons Attribution International 4.0 License.}
  \end{minipage}
  \vspace{5pt}
}
\makeatother

\setcopyright{ifaamas}
\acmConference[AAMAS '25]{Proc.\@ of the 24th International Conference
on Autonomous Agents and Multiagent Systems (AAMAS 2025)}{May 19 -- 23, 2025}
{Detroit, Michigan, USA}{Y.~Vorobeychik, S.~Das, A.~Nowé  (eds.)}
\copyrightyear{2025}
\acmYear{2025}
\acmDOI{}
\acmPrice{}
\acmISBN{}

\usepackage{balance} 

\usepackage{tcolorbox}
\usepackage{enumitem}
\usepackage{todonotes}

\usepackage{listings}
\usepackage{pifont}
\usepackage{float}
\usepackage{tabularx}
\usepackage{fvextra}
\usepackage{fancyvrb}
\usepackage{newtxtext}
\usepackage{newtxmath}
\usepackage{verbatimbox}
\usepackage{etoolbox}

\makeatletter
\preto{\@verbatim}{\topsep=0pt \partopsep=0pt }
\makeatother
\newcommand{\codeexampleone}{{\color{orange} $\text{\{CODE EXAMPLE ONE\}}$}}
\newcommand{\codeexampletwo}{{\color{orange} $\text{\{CODE EXAMPLE TWO\}}$}}
\newcommand{\imone}{{\color{orange} $\text{\{IMPERFECT INFO EXAMPLE ONE\}}$}}
\newcommand{\imtwo}{{\color{orange} $\text{\{IMPERFECT INFO EXAMPLE TWO\}}$}}
\newcommand{\imthree}{{\color{orange} $\text{\{IMPERFECT INFO EXAMPLE THREE\}}$}}
\newcommand{\gamedescription}{{\color{orange} $\text{\{GAME DESCRIPTION\}}$}}
\newcommand{\api}{{\color{orange} $\text{\{API DOCUMENTATION\}}$}}
\newcommand{\imperfect}{{\color{blue} $\text{\{CODE FOR IMPERFECT INFORMATION\}}$}}
\newcommand{\error}{{\color{blue} $\text{\{ERROR MESSAGE\}}$}}
\newcommand{\derror}{{\color{orange} $\text{\{GENERAL GUIDANCE ON ERRORS\}}$}}
\newcommand{\constraints}{{\color{orange} $\text{\{GUIDANCE ON CODE\}}$}}

\usepackage{xspace}
\newcommand{\passALLfive}{\ensuremath{pass\ all\ 5}\xspace}
\newcommand{\passATfive}{\ensuremath{pass@5}\xspace}

\title{From Natural Language to Extensive-Form Game Representations} 

\author{Shilong Deng}
\affiliation{
  \institution{University of Liverpool}
  \city{Liverpool}
  \country{United Kingdom}}
\email{shilong.deng@liverpool.ac.uk}

\author{Yongzhao Wang}
\affiliation{
  \institution{The Alan Turing Institute;}
  \institution{University of Liverpool}
  \city{Liverpool}
  \country{United Kingdom}}
\email{yongzhao.wang@turing.ac.uk}

\author{Rahul Savani}
\affiliation{
\institution{The Alan Turing Institute;}
  \institution{University of Liverpool}
  \city{Liverpool}
  \country{United Kingdom}}
\email{rahul.savani@liverpool.ac.uk}

\begin{abstract}
We introduce a framework for translating game descriptions in natural language into game-theoretic extensive-form representations, leveraging Large Language Models (LLMs) and in-context learning. 
We find that a naive application of in-context learning struggles on this problem, in particular with imperfect information.
To address this, we introduce \textit{GameInterpreter}, a two-stage framework with specialized modules to enhance in-context learning, enabling it to divide and conquer the problem effectively.
In the first stage, we tackle the challenge of imperfect information by developing a module that identifies information sets and the corresponding partial tree structure. 
With this information, the second stage leverages in-context learning alongside a self-debugging module to produce a complete extensive-form game tree represented using pygambit, the Python API of a recognized game-theoretic analysis tool called Gambit.
Using this python representation enables the automation of tasks such as computing Nash equilibria directly from natural language descriptions.
We evaluate the performance of the full framework, as well as its individual components, using various LLMs on games with different levels of strategic complexity. 
Our experimental results show that the framework significantly outperforms baseline approaches in generating accurate extensive-form games, with each module playing a critical role in its success.
\end{abstract}


\begin{document}


\pagestyle{fancy}
\fancyhead{}


\maketitle 

\begin{tcolorbox}
A short version of this paper appears at AAMAS'25~\cite{DWS25}.
This full version includes a detailed appendix with game descriptions
and prompt information.

\medskip
 
The prompts, game descriptions, and also the generated (valid and invalid) extensive-form game
files from our experiments are available for download from the following repository:
\url{https://github.com/zczlsde/GameInterpreter}.

\end{tcolorbox}


\section{Introduction}
Recently, large language models (LLMs) have shown remarkable proficiency in handling complex tasks across various domains, including code generation \cite{chen2021evaluating,li2022competition,fenia2022pangu,roziere2023code} and question answering \cite{tan2023can,li2024flexkbqa,huang2023dsqa}. 
Their success has sparked interest in exploring their potential across an even broader range of applications. 
Within the field of multi-agent systems, a primary research direction focuses on developing LLMs' capabilities for reasoning about \textit{games} and making decisions directly from textual information.
For instance, \citet{fu2023improving} applied LLMs to a bargaining game where LLMs serve as bargaining agents, engaging in price negotiations across several rounds.
In this scenario, a successful bargaining agent must anticipate the behavior and private information of other agents, which requires strong game reasoning skills.

Although there has been initial progress in this area, conducting game-theoretic analysis directly from textual descriptions (such as natural language game descriptions) remains challenging due to the varying degrees of strategic complexity in games, including imperfect information, chance events, and repeated interactions. 
Broadly, there are two technical approaches to tackle this task.
The first approach involves training LLMs specifically to perform game-theoretic analysis. 
This method enables LLMs to conduct analysis directly but often requires extensive training data in games and their descriptions, as well as significant computational resources.
The second approach utilizes LLMs to interpret game descriptions and generate structured representations that can then be analyzed using game-theoretic methods. 
Rather than equipping LLMs with full reasoning abilities, this approach integrates LLMs into the automated reasoning process, which can reduce computational demands.

In this work, we pursue the second approach, presenting \textit{GameInterpreter}, a framework that translates natural language game descriptions into the ``extensive-form''.
An extensive-form game (EFG) is a standard game-theoretic representation for sequential games~\cite{von2021game}.
It is a rooted \textit{tree} with additional information structure called \textit{information sets}, which group together decision \textit{nodes} that are indistinguishable to a player.
Our framework relies on LLMs and in-context learning, where the LLM uses context or examples in the input prompt to complete tasks without the need for fine-tuning or further training.
However, due to the strategic complexities outlined above, directly describing the task in the prompts for in-context learning is insufficient. 
Among these complexities, we particularly emphasize the issue of imperfect information -- that is, where at least one player does not have full knowledge about the current state of the game -- which leads LLMs with naive in-context learning to produce incorrect game representations, as demonstrated in our experiments.

To address this, we take a divide-and-conquer approach using a two-stage process. 
In the first stage, we focus on any imperfect information in the game, by guiding LLMs through examples of dealing with imperfect information (e.g., identifying information sets) and the corresponding partial tree structures. 
With this foundation, the second stage leverages in-context learning to generate the complete EFG for the target game. 
The EFG is created using pygambit, the Python API for the widely used game-theoretic tool Gambit~\cite{savani2024gambit}, which also enables automating tasks such as computing Nash equilibria from natural language descriptions. 
Additionally, we introduce a self-debugging module that returns pygambit error messages to the LLMs, which allows the LLM to correct the errors in its previous answers and helps to ensure that a valid EFG is created.

We assess the performance of our framework, as well as its individual components, across various LLMs, on games featuring differing levels of strategic complexity, covering different numbers of players, degrees of imperfect information, perfect/imperfect recall, and various game tree depths. 
We use two datasets of game descriptions, one newly designed for this paper, and another from a recent paper by others~\citet{mensfelt2024autoformalization}. 
The LLMs we employ are GPT-3.5 \cite{brown2020language}, GPT-4 \cite{achiam2023gpt}, and GPT-4o \cite{openai2024gpt4o}, and we evaluate their ability to generate correct EFG files by incrementally adding modules until the full framework pipeline is assembled. 
Our findings indicate that the full pipeline significantly enhances performance across all LLMs, with the best-performing model successfully solving all test games in our custom dataset. 
Additionally, we confirm that each module significantly contributes to better performance. 
The second dataset of \citet{mensfelt2024autoformalization}
mainly comprises two-player simultaneous-move games, with many different game descriptions for the same underlying bimatrix game; our full pipeline achieves 100\% accuracy on these games, demonstrating robustness of varying game descriptions.
Our main contributions are:
\begin{enumerate}[leftmargin=6mm]
    \item An in-context LLM framework for translating game descriptions in natural language into extensive-form representations;
    \item An imperfect information retrieval module that identifies information sets and the corresponding partial tree structure;
    \item A self-debugging module that ensures the generated code complies with pygambit;
	\item A comprehensive evaluation of our framework, which demonstrate that it 
		significantly outperforms baseline approaches.
\end{enumerate}


\section{Related Work}

\subsection{LLMs with Game Theory}

Many papers have explored the use of LLMs as agents to play games, 
ranging from simple matrix games~\cite{akata2023playing,silva2024large,poje2024effect,lore2023strategic}, to much more complex environments~\cite{mao2023alympics,shi2023cooperation,xu2023exploring}.
\citet{akata2023playing} revealed the different behavioral patterns of LLMs when playing in various types of games.
\citet{shi2023cooperation} examined the ability LLM agents to cooperate in the Avalon game, and developed a memory-based module to facilitate the cooperation.
\citet{xu2023exploring} tested LLM agents in the incomplete information game Werewolf. 
They observed emergent strategic behaviors such as deception during gameplay.
\citet{fan2024can} analyzed the rationality of LLMs as agents, focusing on three specific aspects: building a clear desire, refining beliefs about uncertainty, and taking optimal actions.
\citet{silva2024large} explored if LLMs can be used as an equilibrium solver for games, and 
highlighted the difficulty of this for games with only mixed-strategy equilibria, providing
enhancements to address this.

Besides using LLMs as agents, game-theoretic approaches could be utilized to improve the performance of LLMs.
\citet{gemp2024states} introduced a method that feeds the outputs from game-theoretic methods (e.g., an equilibrium distribution over instructions) to LLM agents in dialogues that can be formed as EFGs.
They demonstrated that the integration with game-theoretic outputs could enhance the LLM generations compared to a baseline LLM that lacks access to game-theoretic supports.
\citet{ma2023red} studied the value alignment problem in LLMs.
They gamified the attacks and counterattacks among LLMs and used equilibrium solutions to improve the level of value alignment of LLMs.
Similarly, \citet{jatova2024employing} framed the generation of toxic content and defence against this as a strategic game between a language model and an adversarial prompt generator, and the equilibrium of the game demonstrated a reduction of harmful outputs.

\subsection{Game Description Translation}
We are aware of three works that directly addresses the task of game description translation \cite{mensfelt2024autoformalization, mensfelt2024autoformalizing,daskalakis2024charting}. 
The earliest one by \citet{mensfelt2024autoformalization} is contemporaneous and independent work with ours.
Rather than using EFGs, for representing games, \citet{mensfelt2024autoformalization} employed logic representations as used by logic programming solvers.
A further key difference between our work and theirs lies in the scope of games analyzed. 
\citet{mensfelt2024autoformalization} focused on simultaneous-move games, with 110 bimatrix games of size 2x2\footnote{A 2x2 bimatrix game corresponds to an EFG with 3 nodes and 4 outcomes (i.e., leaves).}, one bimatrix game of size 3x3, and one sequential game that corresponds to a bimatrix game of size 2x4. In their subsequent work, \citet{mensfelt2024autoformalizing} examined 55 simultaneous-move 2x2 bimatrix games.
In contrast, our work explores more complex scenarios, with multiple sequential moves and complicated (imperfect) information structures.
Both work leverage the ability of LLMs to generate code, which we discuss next. 
The same framework was also adopted in their follow-up work by \citet{mensfelt2024autoformalizing}.
In this follow-up work, they presented a necessary condition for automatically verifying the correctness of the generated game.
In addition to these two works, \citet{daskalakis2024charting} converted the sequential decision-making process described in the game derived from a story to an EFG. 
They achieved this by utilizing LLMs to introduce additional decision nodes, representing alternative choices players could have made in the story. 
Once the EFG is constructed, Gambit is used to compute the Nash equilibrium, providing a prediction of the players' behavior.

More broadly, several studies have used LLMs to translate general texts (i.e., not necessarily game descriptions) in natural language into formal specifications (i.e., required format of a software system or hardware component) \cite{hahn2022formal, zhai2020c2s, leite2024extracting, leong2023translating}. 
\citet{hahn2022formal} examined the ability of fine-tuned language models to convert natural language into formal specifications, which can be used in software verification, theorem proving, and industrial hardware.
\citet{zhai2020c2s} and \citet{leong2023translating} focused on translating textual requirement descriptions into Java formal specifications.
\citet{leite2024extracting} employed LLMs to generate specifications for smart contracts. 

\subsection{LLMs for Code Generation}

Code generation is the process of automatically creating source code using LLMs based on natural language task descriptions. 
Since the introduction of models like Codex~\cite{chen2021evaluating}, Alphacode \cite{li2022competition}, Pangu-Coder~\cite{fenia2022pangu}, and LLaMa Coder~\cite{roziere2023code}, general code generation has made significant advancements with the emergence of models such as LLaMa 3~\cite{dubey2024llama} and GPT-4~\cite{achiam2023gpt}. 
Building on these models, many studies \cite{le2022coderl,liu2023rltf,shojaee2023execution} further improved the performance of code generation through reinforcement learning or self-debugging prompts.
In our work, we leverage the robustness of code generation (with self-debugging) for generating EFG representations in pygambit.


\section{Preliminaries}
\subsection{In-Context Learning}

In-context learning refers to an LLM's ability to learn new information or skills by
observing examples or instructions provided in its input, without any additional training or
fine-tuning~\cite{dong2022survey}.
Suppose we have an LLM, represented by a function $\mathcal{M}$.
It has been trained to predict the next token or word, probabilistically mapping
an input prompt to an output.
An input prompt is a sequence of tokens $z = (z_1, z_2, \dotsc, z_n)$. 
Given $z$, the probabilities for the next token $y$ are then $\mathcal{M}(y|z) = P(y|z_1, z_2, \dotsc, z_n)$.
For in-context learning, consider an input prompt that contains a sequence 
$(Q_i, A_i)$ of examples, which could be in the form of question-answer pairs,
or these could be examples of how to solve examples of a specific task.
We use these as context to predict an answer $A_{n+1}$ for a new question~$Q_{n+1}$, with
our probabilistic answer being $\mathcal{M}(A|(Q_1, A_1),(Q_2, A_2),\dotsc,(Q_n, A_n),Q_{n+1})$.
Importantly, with in-context learning, the model does not adjust its weights,
but instead uses the examples in its the context window to 
to refine its conditional probability for the next token.

\subsection{Extensive-Form Game Representations}\label{sec: EFG}

Extensive-form representations explicitly capture \emph{sequential} decision making, in contrast to the strategic 
form, which condenses the game into a payoff matrix, interpreted as a simultaneous-move
game.
Therefore, EFGs are more expressive than strategic-form games, making them our preferred target
format game translation.
An extensive-form game consists of the following three elements:

\smallskip

\noindent
\textbf{Game tree:} The central element of an EFG is a \emph{rooted, directed tree}. 
Each decision node in the tree is assigned to either one of the players or to a ``chance node,'' which represents nature\footnote{Chance nodes allow random behavior such as card deals in poker.}. Directed edges
are known as moves or actions. A play of the game starts at the root and advances through the tree
as the player that owns the current decision node chooses a move at the node (or a move is drawn
from a discrete probability distribution at a chance node). The play ends when a terminal node
(leaf) of the tree is reached.

\smallskip

\noindent
\textbf{Outcomes and payoffs}: Every terminal node corresponds to an outcome, with an associated
payoff vector which prescribes the payoff for each player under this outcome.

\smallskip

\noindent
\textbf{Information sets}: An information set groups together nodes of a given player, with the
interpretation that a player knows they are at some node in an information set, but not which
one\footnote{Thus, all nodes in the information set must have exactly the same available moves.}.
Information sets allow us to model a player's lack of knowledge of past moves of other players including nature (or the same player in the case of imperfect recall). If all information
sets are singletons the game is said to have perfect information, otherwise it is said to have
imperfect information.

\subsection{Code Generation with Gambit}
\begin{figure}[!htpb]
  \centering
  \includegraphics[width=1.0\linewidth]{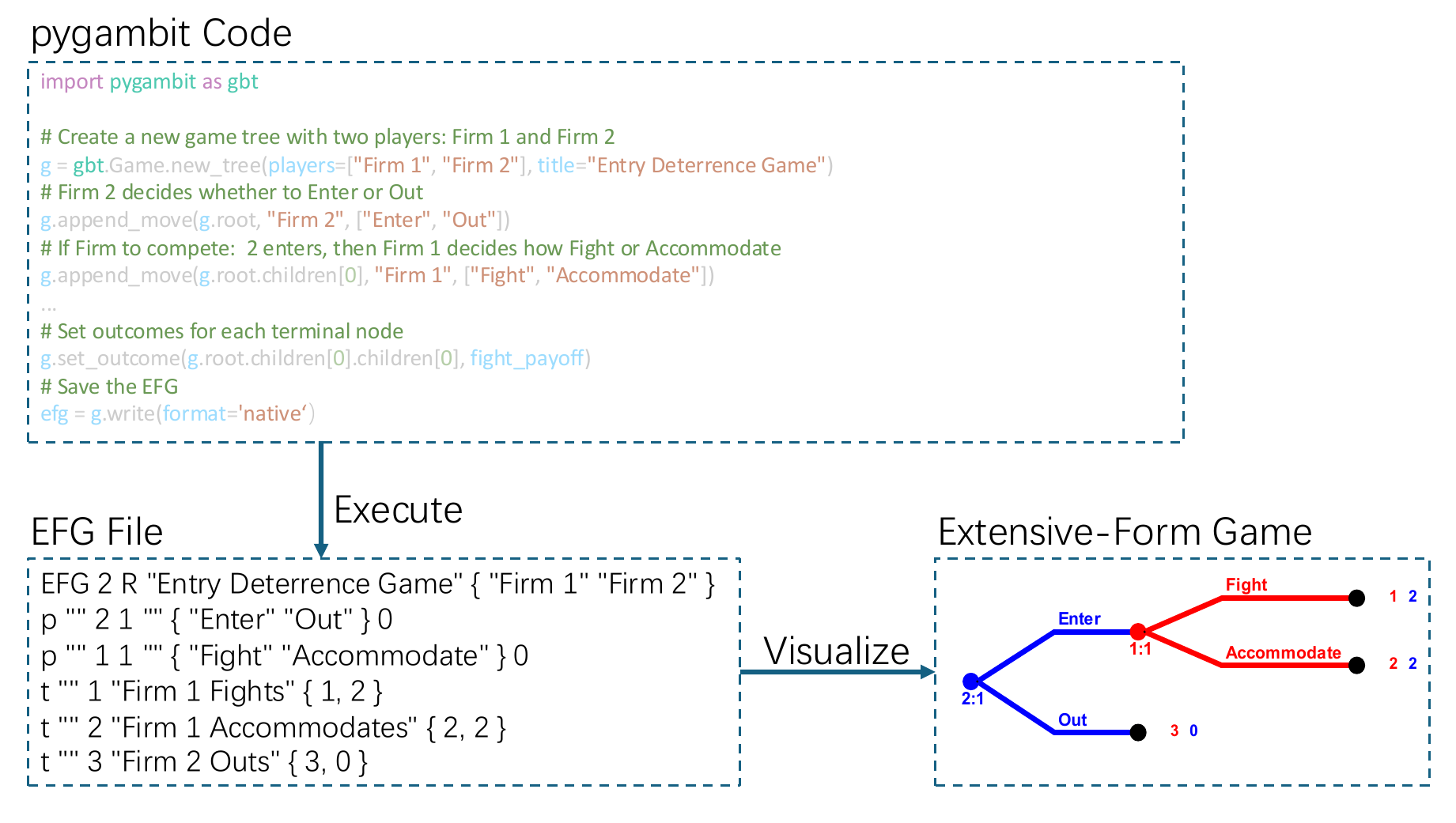}
  \caption{Example of generating and then visualizing an EFG file for a simple Market Entry Game, via pygambit.}
  \label{fig:representation}
  \Description{}
\end{figure}

\textit{Gambit} is a software suite for game-theoretic analysis~\cite{savani2024gambit}.
Gambit can compute game-theoretic solutions such as Nash equilibria and equilibrium refinements
for strategic-form games and EFGs.
It has a Python API known as \textit{pygambit}.
With pygambit, we create an internal representation of an EFG and export it to a standard file format, specifically an .efg file. 
This file can then be analyzed or visualized using Gambit or other compatible software.
Figure~\ref{fig:representation} provides an illustration.

The following pygambit functions for creating and manipulating EFGs are used in 
our guidance examples within our framework:
\begin{enumerate}[leftmargin=7mm]
    \item \textbf{new\_tree()}: create a trivial game tree with one node;
    \item \textbf{append\_move()}: add a move at a terminal node;
    \item \textbf{add\_outcome()}: introduce a new outcome to the game;
    \item \textbf{set\_outcome()}: assign payofs to an outcome;
    \item \textbf{set\_chance\_probs()}: set chance node move probabilities;
    \item \textbf{set\_infoset()}: assign a node to an information set.
\end{enumerate}


\section{The Game Interpreter Framework}

\begin{figure*}[!htpb]
  \centering
  \includegraphics[width=0.8\linewidth]{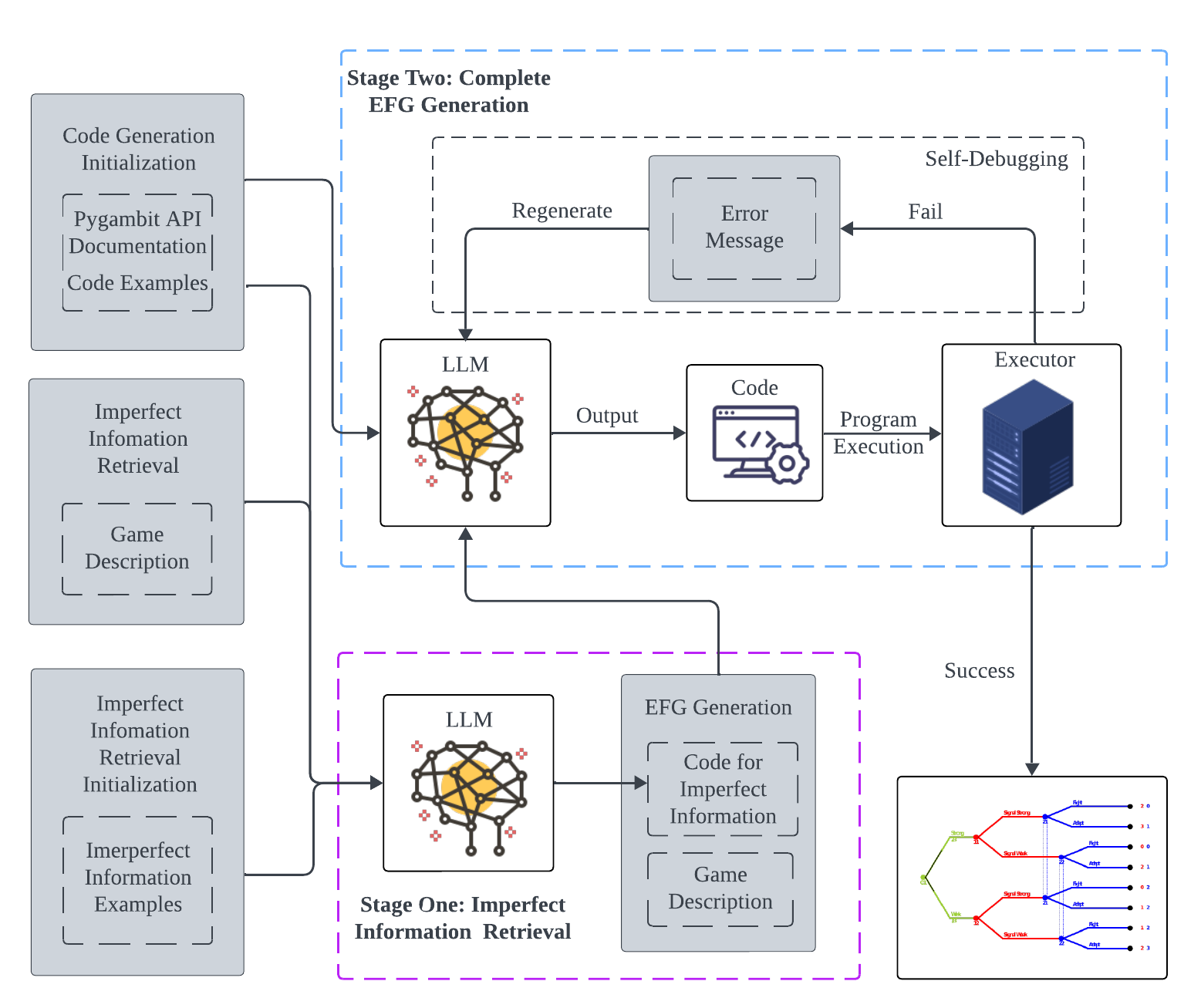}
  \caption{Full GameInterpreter pipeline with all modules included. The rectangles with a gray background represent the prompts given to the LLM and details of these prompts are shown in Table~\ref{tab:prompt}.}
  \label{fig:pipeline}
  \Description{}
\end{figure*}

In Figure~\ref{fig:pipeline}, we present the full GameInterpreter framework for translating natural language game descriptions into EFG files. It involves two stages: imperfect information retrieval and complete EFG generation.
In the first stage, we address the challenge of handling imperfect information by employing in-context learning to identify \emph{non-singleton} information sets and their associated partial tree structures. 
The inputs for this in-context learning, detailed further below, include: general information about the task and the use of the pygambit API, a description of the target game, and instructions with examples for extracting imperfect information.
At this stage, the expected output is a code block. For imperfect information games, the code block (should\footnote{We cannot guarantee the LLM produces what it was asked to, but for brevity in the rest of our description we simply 
say that the output ``includes'' rather that ``should include''.}) include a set of information sets defined by the function \textbf{set\_infoset()}, which groups decision nodes that a player cannot distinguish between, accompanied by reasoning provided in the code comments.
For perfect information games, the output contains only code comments, which includes a concluding statement such as ``there is no need to set any information sets in this game,'' along with reasoning for this conclusion.

\begin{figure}[!htpb]
  \centering
  \includegraphics[width=0.70\linewidth]{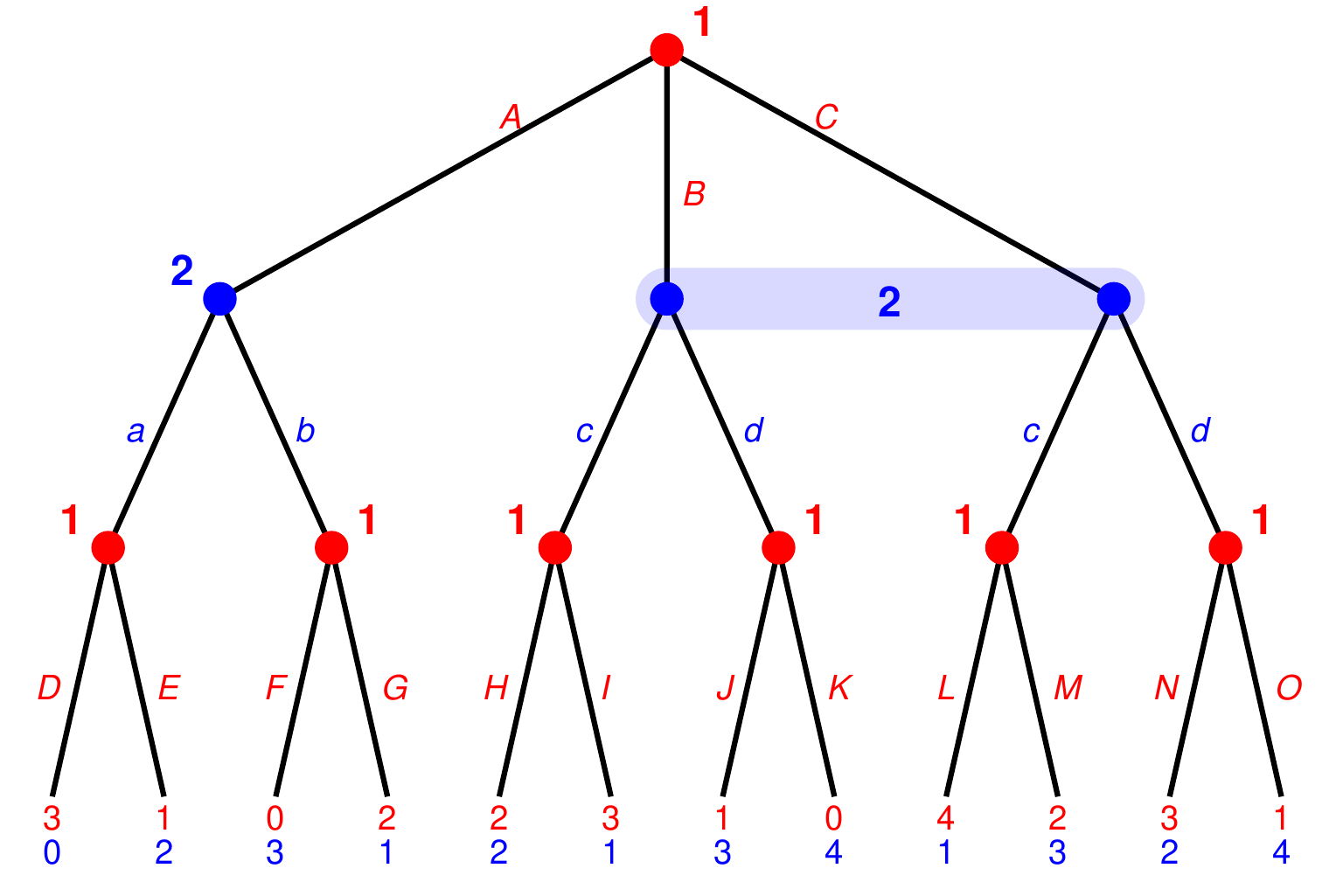}
  \caption{EFG with imperfect information.}
  \label{fig:imperfect_example}
  \Description{}
\end{figure}

Notably, generating these information sets provides insights into the EFG tree structure. 
For example, Figure~\ref{fig:imperfect_example} shows an EFG with two players. 
Player 1 moves first by choosing one of three actions: L, C, or R. 
If either C or R is selected, then player 2 does not know which was selected (i.e., imperfect information).
In contrast, if player~1 chooses action~L, the resulting subgame has perfect information.
After applying our method from stage one, the LLM can separate the perfect information part of the tree from the part with imperfect information.
Then it will only assign nodes to information sets (using \textbf{set\_infoset()}) 
only in the imperfect information part, and it also provides a textual description of the game's tree structure\footnote{Note that the node assignment already reveals insights into the tree structure such as the action that leads to the node in the information set.}.

The second stage employs in-context learning to generate a complete EFG file. 
The inputs for in-context learning in this stage also contain the code generation initialization, instructions for generating the entire EFG, and additionally the output from stage one.
Besides, a self-debugging module is introduced to help the LLM produce runnable pygambit code. 
This module sends any error messages back to the LLM for code revision, providing both a description of the issue and instructions for resolving it, which are essential for fixing mistakes in previous answers. 
This self-debugging approach, also referred to as self-reflection, is discussed in detail by~\citet{shinn2024reflexion}.

Table~\ref{tab:prompt} outlines the prompt templates used for the inputs of in-context learning.
A detailed explanation of each template is provided below. 

\smallskip

\noindent
\textbf{Code Generation Initialization (CGI).}
The CGI is designed to enhance the LLM's understanding of how to use the pygambit library, beyond its prior knowledge, by examining example code. 
We provide two examples, each containing a game description and the corresponding pygambit code for generating the EFG.
We also provide additional guidance through pygambit API documentation.

\smallskip

\noindent
\textbf{Imperfect Information Retrieval Initialization (IIRI).}
This initialization aims to guide the LLMs to extract imperfect information from a game description and use the \textbf{set\_infoset()} function to express the imperfect information.
In the prompt, we present three demonstrations.
Each demonstration includes a game description, the reasoning process for identifying imperfect
information, and the code that groups decision nodes using the \textbf{set\_infoset()} function.

\smallskip

\noindent
\textbf{Imperfect Information Retrieval.}
The prompt for the retrieval process includes specific guidance on the task of extracting
incomplete information from a target game description.
For example, we request that \textbf{set\_infoset()} is used, and also ask the LLM to provide
its reasoning, as in the Chain-of-Thought (CoT) methodology~\cite{wei2022chain}.

\smallskip

\noindent
\textbf{EFG Generation.}
Finally, in stage two, the EFG generation prompt asks the LLM 
to create an EFG file based on the target game description.
For imperfect information games, it explicitly incorporates the output from stage one. 
For perfect information games, it utilizes the conclusion of stage one that no imperfect information is present in the games.
Guidance on common pygambit bugs is provided and,
again, we ask the LLM to give its reasoning as in the CoT approach.

\smallskip

\noindent
\textbf{Self-debugging.}
The prompts are only used if an error is encountered with the LLM's pygambit code output.
The prompts guide the LLMs to correct these errors in a next attempt, and include  
two components: the python interpreter's error message and instructions on addressing common bugs
encountered when using the pygambit library.
To ensure a fair comparison, in all settings without self-debugging, if errors are encountered,
we provide the LLM with a prompt requesting a new response without including any bug-related information or additional
instructions. 
This setup isolates the effect of self-debugging when comparing to settings without it.

\begin{table}[!htpb]
	
\scalebox{0.88}{\begin{tabular}{m{60pt}<{\centering}m{180pt}}\toprule
\textbf{Inputs} & \textbf{Prompts}\\
\midrule
\textbf{Code Generation Initialization} & Given a game description in natural language, you will be asked to generate python code for the Gambit API (pygambit) to construct a corresponding extensive-form game in Gambit.
Here are two examples of how to use pygambit library:

\codeexampleone

\codeexampletwo

Below is the documentation for several relevant functions in the pygambit library:

\api
\\ \midrule
\textbf{Imperfect Information Retrieval Initialization}  & A challenge of this task is to represent 
the imperfect information in the game with pygambit.
Given the game description below, please infer the imperfect information structure in 
the game. Make sure that if there are multiple decision nodes of a player who cannot tell
among these nodes which node they are at, then these nodes are all grouped in the same information set.
In short, an information set belongs to a player and should contain all nodes of that player such that 
the player will know that they are at one of these nodes but they will not know exactly which one they are at.

\imone

\imtwo

\imthree   
   \\ \midrule
  
  \textbf{Imperfect Information Retrieval} & \gamedescription 
  
  You MUST ONLY include the necessary \textbf{set\_infoset()} functions in the Python code block.

    Do NOT include any other code in the code block.
    
    Think step by step and write your reasoning in comments (step-by-step thought process) within the code. \\ \midrule   \midrule

    \textbf{EFG Generation} & 
   
   \gamedescription 
   
  The CODE for representing the imperfect information of the game is as follows:

   \imperfect

   \constraints

  Then, could you write python code to generate the EFG for this game using the pygambit library in the example? 
  
  Let's think step by step and write the reasoning in the code comments. \\  \midrule  \midrule
  \textbf{Error Message} & Your code contains an error. Please review and fix it before trying again. 
  
  \error 
  
  \derror
  \\
	    \bottomrule
\end{tabular}}
 
 \caption{Input prompts of the framework.}
 \label{tab:prompt}
\end{table}


\section{Experimental Setup}
\subsection{Datasets}\label{sec: dataset}

In our experimental evaluation, we used two datasets: a custom dataset created specifically for this study, which focuses on sequential games with a single description provided for each underlying game, and a dataset from \citet{mensfelt2024autoformalization}, which emphasizes bimatrix (simultaneous-move) games and includes multiple descriptions for the same underlying game. 
The latter dataset is particularly useful for assessing the robustness of our method to variations in descriptions.

\smallskip

\noindent
\textbf{New custom dataset:}
Our 18 game descriptions in this dataset, each correspond to a different underlying game.
These games have a range of different strategic complexities, as shown in Table~\ref{tab:dataset},
where we display certain characteristics of these games: binary features such as whether they are
perfect information, or zero-sum; and numeric features such as the maximum depth of the game tree
and the number of players, decision nodes, and leaves. 
The 18 games are chosen to cover classic games from the literature such as Kuhn poker and
Tic-Tac-Toe.
Some of these games were adapted from their standard forms and others were
not not taken from literature or teaching materials, to mitigate the risk that the LLMs had memorized the answers to our requests\footnote{In our experiments, when we ask directly for an .efg file (Setting A below) or pygambit code (Settings B,C,D below) as output, this could
	happen if the LLM had seen this file or code during training. To address this concern, we carefully curated our test set to include many games for which it is highly unlikely that corresponding pygambit code or .efg files were present in the training data.}.
For example, our game ``Nim with five in one pile'' is adapted from an .efg file obtained from the Gambit website. However, we modified the payoffs by switching the game from normal play to misère. 
Additionally, the three games listed as ``Extra'' Games One, Two, and Three in Table~\ref{tab:dataset} were created by us and are not derived from any pre-existing materials available online. 
Also note that games like Colonial Control and Tic-Tac-Toe did not historically have EFG files or pygambit code available on the web, meaning the LLM could not rely on memorization to translate these games.

\smallskip

\noindent
\textbf{Dataset from \citet{mensfelt2024autoformalization}.}
This dataset consists of 112 descriptions in total, which correspond to:
110 bimatrix games of size 2x2, 1 bimatrix game of size 3x3,
and 1 is a sequential game that corresponds to a 2x4 bimatrix game. 
The 110 bimatrix games are based on five classic games: Battle of the Sexes, Hawk-Dove, Matching
Pennies, Prisoner’s Dilemma, and Stag Hunt. 
Each classic game is presented in two formats: one with numerical payoffs and one without. 
For each format, there are a total of 11 descriptions, including one standard explanation from the game theory literature and 10 variations generated using GPT-4o, which reinterpret these games as diverse real-life scenarios.

\begin{table*}[!htpb] 
	
	\centering
	\setlength{\tabcolsep}{4mm}{
	\scalebox{0.9}{\begin{tabular}{lcccccc}
		\toprule
		\textbf{Game Names} & \multicolumn{6}{c}{\textbf{Game Features}} \\
		\cmidrule(lr){2-7}
		 & \textbf{Perfect Info} & \textbf{Zero-Sum}  & \textbf{Max Depth} & \textbf{$\#$Players} & \textbf{$\#$Nodes} & \textbf{$\#$Leaves} \\
		\midrule
		\textbf{A Three-Player Game} & \textcolor{BrickRed}{\ding{55}} & \textcolor{ForestGreen}{\ding{51}} & 4 & 3 & 7 & 8\\
            \midrule
		\textbf{An Imperfect Recall Game} & \textcolor{BrickRed}{\ding{55}} & \textcolor{ForestGreen}{\ding{51}} & 3 & 2 & 7 & 8\\
            \midrule
		\textbf{Absent-Minded Driver} & \textcolor{BrickRed}{\ding{55}} & \textcolor{BrickRed}{\ding{55}} & 2 & 1 & 2 & 3\\
            \midrule
		\textbf{Bach or Stravinsky} & \textcolor{BrickRed}{\ding{55}} & \textcolor{BrickRed}{\ding{55}} & 2 & 2 & 3 & 4\\
		\midrule
		\textbf{Bagwell} & \textcolor{BrickRed}{\ding{55}} & \textcolor{BrickRed}{\ding{55}} & 3 & 2 & 7 & 8\\
             \midrule
		\textbf{Kuhn Poker} & \textcolor{BrickRed}{\ding{55}} & \textcolor{ForestGreen}{\ding{51}} & 4 & 2 & 25 & 30 \\
        \midrule
        \textbf{Extra Game One} &  \textcolor{BrickRed}{\ding{55}} & \textcolor{BrickRed}{\ding{55}} & 5 & 2 & 16 & 21 \\
        \midrule
        \textbf{Extra Game Two} & \textcolor{BrickRed}{\ding{55}} & \textcolor{BrickRed}{\ding{55}} & 5 & 3 & 22 & 24 \\
        \midrule
		\textbf{Market Signalling Game} & \textcolor{BrickRed}{\ding{55}} & \textcolor{BrickRed}{\ding{55}} & 3 & 2 & 7 & 8\\
        \midrule
		\textbf{Nuclear Crisis} & \textcolor{BrickRed}{\ding{55}} & \textcolor{BrickRed}{\ding{55}} & 4 & 2 & 5 & 6\\
        \midrule
		\textbf{Rock, Paper, Scissors} & \textcolor{BrickRed}{\ding{55}} & \textcolor{ForestGreen}{\ding{51}} & 2 & 2 & 4 & 9\\
        \midrule
        \midrule
        \textbf{Centipede} & \textcolor{ForestGreen}{\ding{51}} & \textcolor{BrickRed}{\ding{55}} & 4 & 2 & 4 & 5\\
        \midrule
		\textbf{Colonial Control} & \textcolor{ForestGreen}{\ding{51}} & \textcolor{BrickRed}{\ding{55}} &  3 & 2 & 4 & 5\\
        \midrule
        \textbf{Extra Game Three} & \textcolor{ForestGreen}{\ding{51}} & \textcolor{ForestGreen}{\ding{51}} & 4 & 2 & 17 & 24 \\
        \midrule
		\textbf{Market Entry Model}  & \textcolor{ForestGreen}{\ding{51}}  & \textcolor{BrickRed}{\ding{55}} & 2 & 2 & 2 & 3\\
        \midrule
          \textbf{Nim (with five in one pile)} & \textcolor{ForestGreen}{\ding{51}} & \textcolor{ForestGreen}{\ding{51}} &  5 & 2 & 12 & 8\\
          \midrule
		\textbf{Simple Bargaining Game} & \textcolor{ForestGreen}{\ding{51}} & \textcolor{BrickRed}{\ding{55}} & 5 & 2 & 5 & 3\\
		\midrule
		\textbf{Tic-Tac-Toe} & \textcolor{ForestGreen}{\ding{51}} & 
        \textcolor{ForestGreen}{\ding{51}}  & 3 & 2 & 5 & 5\\
		\bottomrule
	\end{tabular}}}
 \caption{The games in our custom dataset used in our evaluation, along with their features.}
 \label{tab:dataset}
\end{table*}

\subsection{Baselines}\label{sec: baselines}
\begin{figure}[!htpb]
  \centering
  \includegraphics[width=0.87\linewidth]{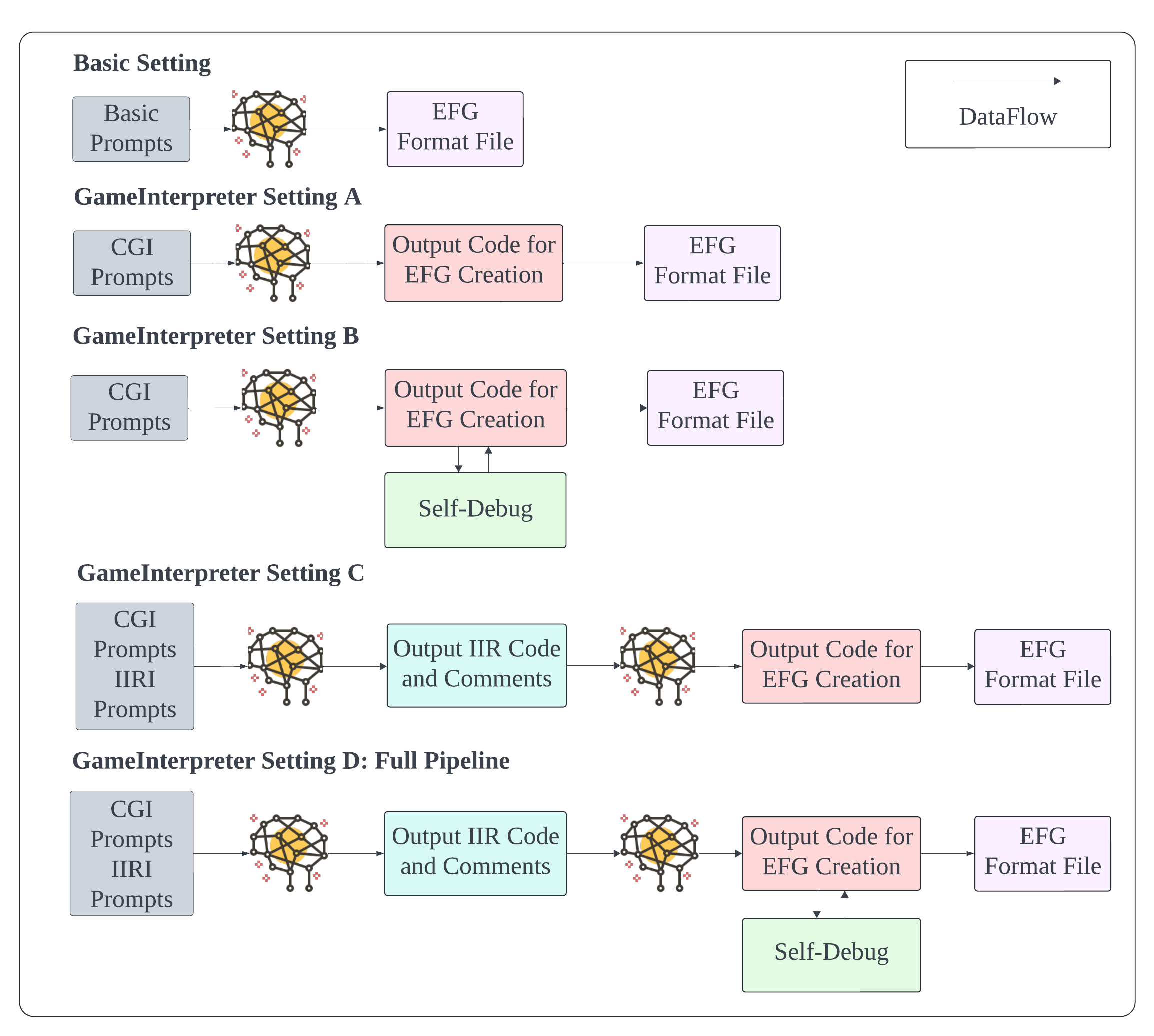}
  \caption{The five approaches we evaluated.}
  \label{fig:settings}
  \Description{}
\end{figure}

We evaluate the effectiveness of each module in our framework by experimenting with various settings, both including and excluding the modules.
In our most basic setting, we task the LLM with generating an EFG file directly based on a game description, without requiring it to utilize the pygambit API.
We then investigate four settings that require the pygambit API:
\begin{itemize}[leftmargin=6mm]
    \item \textbf{Setting A}: The setting that utilizes minimal EFG generation prompts in Table~\ref{tab:prompt}, without incorporating additional information for imperfect information retrieval;
    \item \textbf{Setting B}: Setting A with self-debugging; 
    \item \textbf{Setting C}: Setting A with imperfect information retrieval;
    \item \textbf{Setting D (Full Pipeline)}: Setting A with both self-debugging and imperfect information retrieval.
\end{itemize}
Figure~\ref{fig:settings} provides an illustration. 
For each of these settings, we combine them with three different LLMs.
Additionally, we compare our setting D, the full pipeline, with the approach by
\citet{mensfelt2024autoformalization}, which uses logic programming to represent
target games.

\subsection{Evaluation}\label{sec: evaluation}
When translating game descriptions into extensive-form representations, a key task is to ensure that the generated EFG is \textit{consistent} with the description. Formally, a game description corresponds to a family of EFGs that are consistent with that description; this family can vary in size, and could possibly even by infinite, depending on the specificity of the game details provided. 
For instance, if a game description includes inequalities or relationships between payoffs but lacks precise values, many payoff assignments may be consistent with it, with different corresponding EFGs. 
Similarly, for descriptions of simultaneous-move games (e.g., bimatrix games), a consistent EFG could depict either of the two
players acting first, with imperfect information modeling the simultaneity of their choices.

For checking the consistency, we manually assess whether the game tree, information sets, and payoffs in the generated EFG 
are consistent with the target game description.
This requires firstly checking the generated tree structure, including which
players move at which nodes, starting from the root, along with how the actions at the nodes lead 
to moves of other players; this is informed by the labels for moves that are provided in the generated
EFG files.
For information sets, we ensure that nodes indistinguishable to a given player are grouped within the same information set. 
Finally, for payoffs, if specific values are provided in the game description, we check that the generated payoffs 
at the appropriate terminal nodes (outcomes) match these values. 
If the game description only implies relative payoff strengths, we verify that the payoffs are consistent with the implied constraints.

We use $pass@k$ (the solve rate given $k$ ``samples'', that is, independently generated EFG files), as proposed by~\citet{chen2021evaluating}, to measure the success of our translations for a specific target game description. 
Specifically, we execute GameInterpreter $k=5$ times in our experiments, 
and we record the total number of generated EFGs, $s \in [1,\ldots,5]$, that are consistent with the target game description.
As described in more detail in Section~\ref{sec:results}, in our experimental analysis, we distinguish between the
case where at least one sample was correct (i.e., $s \ge 1$), so we ``\passATfive'',
and the case where all samples were correct (i.e., $s = 5$), referred to as ``\passALLfive'', which is stricter than \passATfive.
While improvements under \passALLfive are more desirable, we also evaluate \passATfive to highlight any 
relative merits of different settings we investigate, particularly in cases where the strictness of \passALLfive 
might hide differences in performance.


\subsection{Parameters}
We use the OpenAI API to access various LLMs, in particular: gpt-4-0125-preview, gpt-4o, and gpt-3.5-turbo. 
All of these models have two key hyperparameters that relate to how the next tokens are chosen, namely
the temperature and $p$ threshold for top-$p$ sampling; both take values in $[0,1]$.
In top-$p$ sampling~\cite{HoltzmanBDFC20}, also known as nucleus sampling, the threshold $p$ is used to restrict sampling of 
the next token to only the smallest set of most-likely candidates whose cumulative probability exceeds $p$.
We set $p$ as 1 (i.e., we do not restrict the next tokens at all).
We set the temperature of LLMs to $0$, which minimizes the amount of randomness in the chosen tokens (a choice of $1$ would maximize it);
note that setting the temperature to $0$ makes the output as deterministic as possible, but, even with a fixed prompt, the output of these LLMs 
still often varies in repeat trials with the temperature set to $0$, which has been attributed to 
issues like multi-GPU inference with varying GPU clock times. 
When comparing our method to the work by \citet{mensfelt2024autoformalization}, we match their experimental settings and adjust the temperature to $1$.
The maximal number of attempts for the self-debugging module is set to $3$.

\section{Experimental Results}
\label{sec:results}

\subsection{Overview}

Table~\ref{tab:result_all} displays the performance of the baselines discussed in Section~\ref{sec:
baselines}, across the 18 games of our custom dataset (see Table~\ref{tab:dataset}). 
A grey tick indicates that between 1 and 4 of the 5 generated samples were successfully solved, a
red cross means none were solved, and a green tick indicates that all 5 samples were solved.
Thus a grey tick indicates a \passATfive and a green tick indicates a \passALLfive (which is by definition
also a \passATfive).

We observed that for all LLMs, setting D, the full pipeline, outperformed its counterparts (the
basic setting, and settings A, B, and C). 
Among the LLMs, GPT-4o achieved the highest performance, succeeding for \passATfive for all 18 games, while GPT-4 also performed
well, succeeding on 15 of the games, and failing only Kuhn poker, Nim, and Extra Game Two, which we note are three
of the largest games that we considered.

\begin{table*}[!htpb] 
    \centering
    \setlength{\tabcolsep}{1.0mm}{
    \scalebox{0.87}{\begin{tabular}{lccccccccccccccc}
        \toprule
        \textbf{Games} 
        & \multicolumn{3}{c}{\textbf{Basic Setting}} & \multicolumn{3}{c}{\textbf{Setting A}} & \multicolumn{3}{c}{\textbf{Setting B}} & \multicolumn{3}{c}{\textbf{Setting C}} & \multicolumn{3}{c}{\textbf{Setting D}}\\
        \cmidrule(lr){2-4}\cmidrule(lr){5-7}\cmidrule(lr){8-10}\cmidrule(lr){11-13}\cmidrule(lr){14-16}
        & 3.5 & 4 & 4o & 3.5 & 4 & 4o & 3.5 & 4 & 4o & 3.5 & 4 & 4o & 3.5 & 4 & 4o \\
        \midrule
        \textbf{A Three-Player Game} & \textcolor{BrickRed}{\ding{55}} & \textcolor{BrickRed}{\ding{55}} &  \textcolor{BrickRed}{\ding{55}} & \textcolor{BrickRed}{\ding{55}} & \textcolor{BrickRed}{\ding{55}} & \textcolor{BrickRed}{\ding{55}} & \textcolor{BrickRed}{\ding{55}} & \textcolor{BrickRed}{\ding{55}}  & \textcolor{BrickRed}{\ding{55}} & \textcolor{BrickRed}{\ding{55}}& \quad \textcolor{gray}{\ding{51}[4]} & \quad \textcolor{gray}{\ding{51}[2]} & \textcolor{BrickRed}{\ding{55}} & \quad \textcolor{gray}{\ding{51}[4]} & \quad \textcolor{gray}{\ding{51}[2]} \\
        \textbf{An Imperfect Recall Game}  & \textcolor{BrickRed}{\ding{55}} & \textcolor{BrickRed}{\ding{55}} &\textcolor{BrickRed}{\ding{55}}  & \textcolor{BrickRed}{\ding{55}} & \textcolor{BrickRed}{\ding{55}} & \textcolor{BrickRed}{\ding{55}} & \textcolor{BrickRed}{\ding{55}} & \textcolor{BrickRed}{\ding{55}}  &  \textcolor{BrickRed}{\ding{55}}  & \textcolor{BrickRed}{\ding{55}} &  \quad \textcolor{gray}{\ding{51}[4]}   &  \quad \textcolor{gray}{\ding{51}[2]}  & \textcolor{BrickRed}{\ding{55}} &  \quad \textcolor{gray}{\ding{51}[4]}  &  \quad \textcolor{gray}{\ding{51}[4]}  \\
        \textbf{Absent-Minded Driver} & \textcolor{BrickRed}{\ding{55}} & \textcolor{BrickRed}{\ding{55}} & \textcolor{BrickRed}{\ding{55}} & \textcolor{BrickRed}{\ding{55}} & \quad \textcolor{gray}{\ding{51}[1]} & \textcolor{BrickRed}{\ding{55}} & \textcolor{BrickRed}{\ding{55}} & \quad \textcolor{gray}{\ding{51}[2]} & \textcolor{BrickRed}{\ding{55}} & \textcolor{BrickRed}{\ding{55}}& \quad \textcolor{gray}{\ding{51}[4]}& \textcolor{BrickRed}{\ding{55}} &  \textcolor{BrickRed}{\ding{55}} & \quad \textcolor{gray}{\ding{51}[4]} & \quad \textcolor{gray}{\ding{51}[2]} \\
        \textbf{Bach or Stravinsky}  & \textcolor{BrickRed}{\ding{55}} & \textcolor{BrickRed}{\ding{55}} & \textcolor{BrickRed}{\ding{55}} & \textcolor{BrickRed}{\ding{55}} & \textcolor{BrickRed}{\ding{55}}  & \textcolor{BrickRed}{\ding{55}} & \quad \textcolor{gray}{\ding{51}[1]} & \textcolor{BrickRed}{\ding{55}} &\textcolor{BrickRed}{\ding{55}}  & \textcolor{BrickRed}{\ding{55}} & \quad \textcolor{gray}{\ding{51}[4]}& \textcolor{ForestGreen}{\ding{51}} & \quad \textcolor{gray}{\ding{51}[1]} & \quad \textcolor{gray}{\ding{51}[4]} & \textcolor{ForestGreen}{\ding{51}} \\
        \textbf{Bagwell} & \textcolor{BrickRed}{\ding{55}} & \textcolor{BrickRed}{\ding{55}} & \textcolor{BrickRed}{\ding{55}}  & \textcolor{BrickRed}{\ding{55}} & \textcolor{BrickRed}{\ding{55}} & \quad \textcolor{gray}{\ding{51}[4]} & \textcolor{BrickRed}{\ding{55}} & \textcolor{BrickRed}{\ding{55}} & \quad \textcolor{gray}{\ding{51}[4]} & \textcolor{BrickRed}{\ding{55}} & \quad \textcolor{gray}{\ding{51}[1]} & \quad \textcolor{gray}{\ding{51}[4]} & \textcolor{BrickRed}{\ding{55}} & \quad \textcolor{gray}{\ding{51}[1]} & \quad \textcolor{gray}{\ding{51}[4]} \\
        \textbf{Kuhn Poker} & \textcolor{BrickRed}{\ding{55}} & \textcolor{BrickRed}{\ding{55}} & \textcolor{BrickRed}{\ding{55}}  & \textcolor{BrickRed}{\ding{55}} & \textcolor{BrickRed}{\ding{55}} & \textcolor{BrickRed}{\ding{55}} & \textcolor{BrickRed}{\ding{55}} & \textcolor{BrickRed}{\ding{55}} & \textcolor{BrickRed}{\ding{55}}  & \textcolor{BrickRed}{\ding{55}} & \textcolor{BrickRed}{\ding{55}} & \quad \textcolor{gray}{\ding{51}[2]}& \textcolor{BrickRed}{\ding{55}} & \textcolor{BrickRed}{\ding{55}} & \quad \textcolor{gray}{\ding{51}[2]}\\
        \textbf{Extra Game One} & \textcolor{BrickRed}{\ding{55}} & \textcolor{BrickRed}{\ding{55}} & \textcolor{BrickRed}{\ding{55}} & \textcolor{BrickRed}{\ding{55}} & \textcolor{BrickRed}{\ding{55}} & \textcolor{BrickRed}{\ding{55}} & \textcolor{BrickRed}{\ding{55}} & \textcolor{BrickRed}{\ding{55}} &  \textcolor{BrickRed}{\ding{55}} & \textcolor{BrickRed}{\ding{55}} & \quad \textcolor{gray}{\ding{51}[3]} & \quad \textcolor{gray}{\ding{51}[4]} & \textcolor{BrickRed}{\ding{55}} & \quad \textcolor{gray}{\ding{51}[4]} &  \quad \textcolor{gray}{\ding{51}[4]}\\
        \textbf{Extra Game Two} & \textcolor{BrickRed}{\ding{55}} & \textcolor{BrickRed}{\ding{55}} & \textcolor{BrickRed}{\ding{55}} & \textcolor{BrickRed}{\ding{55}} & \textcolor{BrickRed}{\ding{55}} & \textcolor{BrickRed}{\ding{55}} & \textcolor{BrickRed}{\ding{55}} & \textcolor{BrickRed}{\ding{55}} & \textcolor{BrickRed}{\ding{55}} & \textcolor{BrickRed}{\ding{55}} & \textcolor{BrickRed}{\ding{55}} & \textcolor{ForestGreen}{\ding{51}} & \textcolor{BrickRed}{\ding{55}} & \textcolor{BrickRed}{\ding{55}} & \textcolor{ForestGreen}{\ding{51}} \\
        \textbf{Market Signalling Game} & \textcolor{BrickRed}{\ding{55}} & \textcolor{BrickRed}{\ding{55}} & \textcolor{BrickRed}{\ding{55}}  & \textcolor{BrickRed}{\ding{55}} & \textcolor{BrickRed}{\ding{55}} & \textcolor{ForestGreen}{\ding{51}} & \textcolor{BrickRed}{\ding{55}} & \textcolor{BrickRed}{\ding{55}} & \textcolor{ForestGreen}{\ding{51}} & \textcolor{BrickRed}{\ding{55}}& \quad \textcolor{gray}{\ding{51}[2]}& \quad \textcolor{gray}{\ding{51}[4]}&\textcolor{BrickRed}{\ding{55}} & \quad \textcolor{gray}{\ding{51}[2]} & \textcolor{ForestGreen}{\ding{51}} \\
        \textbf{Nuclear Crisis}  & \textcolor{BrickRed}{\ding{55}} & \textcolor{BrickRed}{\ding{55}} & \textcolor{BrickRed}{\ding{55}} & \textcolor{BrickRed}{\ding{55}} &  \textcolor{BrickRed}{\ding{55}} & \textcolor{BrickRed}{\ding{55}} & \textcolor{BrickRed}{\ding{55}} & \textcolor{BrickRed}{\ding{55}} & \textcolor{BrickRed}{\ding{55}} & \textcolor{BrickRed}{\ding{55}}& \quad \textcolor{gray}{\ding{51}[2]}& \quad \textcolor{gray}{\ding{51}[2]}& \textcolor{BrickRed}{\ding{55}}& \quad \textcolor{gray}{\ding{51}[2]} & \quad \textcolor{gray}{\ding{51}[2]} \\
        \textbf{Rock, Paper, Scissors} & \textcolor{BrickRed}{\ding{55}} & \textcolor{BrickRed}{\ding{55}} &\textcolor{BrickRed}{\ding{55}}  & \textcolor{BrickRed}{\ding{55}} & \textcolor{BrickRed}{\ding{55}} & \textcolor{BrickRed}{\ding{55}} & \textcolor{BrickRed}{\ding{55}} & \textcolor{BrickRed}{\ding{55}} & \textcolor{BrickRed}{\ding{55}} & \textcolor{BrickRed}{\ding{55}} & \quad \textcolor{gray}{\ding{51}[2]} & \quad \textcolor{gray}{\ding{51}[3]} & \textcolor{BrickRed}{\ding{55}} & \quad \textcolor{gray}{\ding{51}[2]} & \textcolor{ForestGreen}{\ding{51}} \\
        \hline
        \hline
        \textbf{Centipede} & \textcolor{BrickRed}{\ding{55}} & \textcolor{BrickRed}{\ding{55}} &  \textcolor{BrickRed}{\ding{55}} & \textcolor{BrickRed}{\ding{55}} & \quad \textcolor{gray}{\ding{51}[2]}  & \textcolor{ForestGreen}{\ding{51}}& \textcolor{BrickRed}{\ding{55}} & \quad \textcolor{gray}{\ding{51}[4]} & \textcolor{ForestGreen}{\ding{51}} &  \textcolor{BrickRed}{\ding{55}}& \quad \textcolor{gray}{\ding{51}[4]}& \textcolor{ForestGreen}{\ding{51}}&\textcolor{BrickRed}{\ding{55}}  & \textcolor{ForestGreen}{\ding{51}} & \textcolor{ForestGreen}{\ding{51}} \\
        \textbf{Colonial Control} & \textcolor{BrickRed}{\ding{55}} &  \textcolor{ForestGreen}{\ding{51}} & \textcolor{ForestGreen}{\ding{51}} & \textcolor{BrickRed}{\ding{55}} & \textcolor{ForestGreen}{\ding{51}} & \textcolor{ForestGreen}{\ding{51}} & \textcolor{BrickRed}{\ding{55}}  & \textcolor{ForestGreen}{\ding{51}} & \textcolor{ForestGreen}{\ding{51}} & \textcolor{BrickRed}{\ding{55}}& \textcolor{ForestGreen}{\ding{51}}& \textcolor{ForestGreen}{\ding{51}}&\textcolor{BrickRed}{\ding{55}} & \textcolor{ForestGreen}{\ding{51}} & \textcolor{ForestGreen}{\ding{51}} \\
        \textbf{Extra Game Three} & \textcolor{BrickRed}{\ding{55}} & \quad \textcolor{gray}{\ding{51}[1]} & \textcolor{BrickRed}{\ding{55}} & \textcolor{BrickRed}{\ding{55}} & \textcolor{ForestGreen}{\ding{51}} & \quad \textcolor{gray}{\ding{51}[2]} & \textcolor{BrickRed}{\ding{55}} &  \textcolor{ForestGreen}{\ding{51}}  & \quad \textcolor{gray}{\ding{51}[2]} & \textcolor{BrickRed}{\ding{55}} & \quad \textcolor{gray}{\ding{51}[3]} & \textcolor{ForestGreen}{\ding{51}} & \textcolor{BrickRed}{\ding{55}} & \textcolor{ForestGreen}{\ding{51}}  &  \textcolor{ForestGreen}{\ding{51}}\\
        \textbf{Market Entry Model} & \textcolor{BrickRed}{\ding{55}} & \quad \textcolor{gray}{\ding{51}[2]} & \quad \textcolor{gray}{\ding{51}[2]} & \textcolor{BrickRed}{\ding{55}} & \textcolor{ForestGreen}{\ding{51}} & \textcolor{ForestGreen}{\ding{51}} & \textcolor{BrickRed}{\ding{55}} & \textcolor{ForestGreen}{\ding{51}}  & \textcolor{ForestGreen}{\ding{51}} & \textcolor{BrickRed}{\ding{55}} & \textcolor{ForestGreen} {\ding{51}}& \textcolor{ForestGreen} {\ding{51}}& \textcolor{BrickRed}{\ding{55}}& \textcolor{ForestGreen}{\ding{51}} &  \textcolor{ForestGreen}{\ding{51}}\\
        \textbf{Nim (with five in one pile)} & \textcolor{BrickRed}{\ding{55}} & \textcolor{BrickRed}{\ding{55}} & \textcolor{BrickRed}{\ding{55}} & \textcolor{BrickRed}{\ding{55}} & \textcolor{BrickRed}{\ding{55}} & \quad \textcolor{gray}{\ding{51}[4]} & \textcolor{BrickRed}{\ding{55}}  & \textcolor{BrickRed}{\ding{55}} & \quad \textcolor{gray}{\ding{51}[4]} & \textcolor{BrickRed}{\ding{55}}&  \textcolor{BrickRed}{\ding{55}}&  \textcolor{ForestGreen}{\ding{51}}& \textcolor{BrickRed}{\ding{55}} & \textcolor{BrickRed}{\ding{55}} & \textcolor{ForestGreen}{\ding{51}} \\
        \textbf{Simple Bargaining Game} & \textcolor{BrickRed}{\ding{55}} & \textcolor{BrickRed}{\ding{55}} & \quad \textcolor{gray}{\ding{51}[3]} & \textcolor{BrickRed}{\ding{55}} & \textcolor{ForestGreen}{\ding{51}} & \textcolor{ForestGreen}{\ding{51}} & \textcolor{BrickRed}{\ding{55}} & \textcolor{ForestGreen}{\ding{51}}  & \textcolor{ForestGreen}{\ding{51}} & \textcolor{BrickRed}{\ding{55}} & \quad \textcolor{gray}{\ding{51}[3]}& \textcolor{ForestGreen}{\ding{51}}& \textcolor{BrickRed}{\ding{55}}  &\textcolor{ForestGreen}{\ding{51}}  & \textcolor{ForestGreen}{\ding{51}} \\
        \textbf{Tic-tac-toe} & \textcolor{BrickRed}{\ding{55}} & \textcolor{ForestGreen}{\ding{51}} & \textcolor{ForestGreen}{\ding{51}} & \textcolor{BrickRed}{\ding{55}} & \textcolor{ForestGreen}{\ding{51}} & \textcolor{ForestGreen}{\ding{51}} & \textcolor{BrickRed}{\ding{55}} & \textcolor{ForestGreen}{\ding{51}} & \textcolor{ForestGreen}{\ding{51}}  & \textcolor{BrickRed}{\ding{55}} & \quad \textcolor{gray}{\ding{51}[4]} & \textcolor{ForestGreen}{\ding{51}} & \textcolor{BrickRed}{\ding{55}} & \textcolor{ForestGreen}{\ding{51}} & \textcolor{ForestGreen}{\ding{51}} \\
        \bottomrule
    \end{tabular}}}
	\caption{Results of EFG file generation under various settings. A red cross indicates
	that none of the generated samples passed; a green tick shows that all five generated samples
passed; a grey tick signifies that between 1 and 4 generated samples passed, with the
exact number of successes shown in brackets. Perfect information games (top) and imperfect
information games (bottom) are separated.}

    \label{tab:result_all}
\end{table*}

In the basic setting, as well as in settings A and B, we found that imperfect information games, like ``A Three-Player Game'', are challenging to solve. 
This motivates our approach using a first stage for imperfect information retrieval. 
Their poor performance shows that the LLMs had not effectively memorized the solutions for our custom dataset.

Across all settings, GPT-3.5 underperforms compared to GPT-4 and GPT-4o, and it does not benefit
from the addition of imperfect information retrieval and self-debugging.
Ultimately, these results show that setting D, which represents the full pipeline of our method,
yields the best performance across all configurations.

\begin{table}[h!]
\centering
\resizebox{\linewidth}{!}{
\begin{tabular}{lccccccc}
\toprule
\textbf{Metrics} & \multicolumn{2}{c}{\textbf{GPT-3.5}} & \multicolumn{2}{c}{\textbf{GPT-4}} & \multicolumn{2}{c}{\textbf{GPT-4o}}\\ 
 \cmidrule(lr){2-3} \cmidrule(lr){4-5}  \cmidrule(lr){6-7}

 & Setting C & Setting D & Setting C & Setting D & Setting C & Setting D \\ 

\midrule 
\passATfive     & 0 & 1 & 15 & 15 & 17 & 18 \\ 
\passALLfive & 0 & 0 & 2  & 6  & 9  & 11 \\ 

\bottomrule

\end{tabular}}
\caption{The number of games passed after adding self-debugging, evaluated under \passATfive and \passALLfive.}
\label{tab: self-debugging}
\end{table}

\subsection{Performance of Self-Debugging}

In Table~\ref{tab: self-debugging}, we present the \passATfive and stricter
\passALLfive metrics (extracted from Table~\ref{tab:result_all}), after incorporating the self-debugging module for each LLM. 
From Figure~\ref{tab: self-debugging}, we observed that under \passATfive, both GPT-3.5 and GPT-4o
show improvements, while GPT-4 remains unchanged. 
Under \passALLfive, both GPT-4 and GPT-4o improve, whereas GPT-3.5 fails in all games with or
without self-debugging, with the performance constrained by the limitations of the LLM itself.
These findings show that the self-debugging module contributes to an overall enhancement in the
framework's performance.

\begin{table*}[h!]
\centering
\scalebox{0.8}{
\begin{tabular}{llccccccc}
\toprule

\textbf{Game Types} & \textbf{Metrics} & \multicolumn{2}{c}{\textbf{GPT-3.5}} & \multicolumn{2}{c}{\textbf{GPT-4}} & \multicolumn{2}{c}{\textbf{GPT-4o}}\\ 
\cmidrule(lr){3-4} \cmidrule(lr){5-6} \cmidrule(lr){7-8}

& & Setting B & Setting D & Setting B & Setting D & Setting B & Setting D \\

\midrule

\textbf{Imperfect info (11 games)} &\passATfive      & 1 & 1 & 1 & 9 & 2 & 11 \\ 
                                   &\passALLfive  & 0 & 0 & 0 & 0 & 1 & 4 \\ 

\midrule
\midrule

\textbf{Perfect info (7 games)} &\passATfive       & 0 & 0 & 6 & 6 & 7 & 7 \\ 
                                &\passALLfive   & 0 & 0 & 5 & 6 & 5 & 7 \\ 

\midrule
\midrule 

\textbf{All games (18 games)} &\passATfive       & 1 & 1 & 7 & 15 & 9 & 18 \\ 
                              &\passALLfive   & 0 & 0 & 5 & 6  & 6  & 12 \\ 

\bottomrule

\end{tabular}
}
\caption{The number of games passed after adding imperfect information retrieval, evaluated under \passATfive  and \passALLfive.}
\label{tab:IIR}
\end{table*}

\subsection{Performance of Imperfect Info Retrieval}
In Table~\ref{tab:IIR}, we examine the impact of the imperfect information retrieval module by comparing setting B with the full pipeline, 
under the \passATfive and stricter \passALLfive metrics (extracted from Table~\ref{tab:result_all}).
We distinguish between imperfect information games and perfect information games to analyze the module's effect on each category.
Table~\ref{tab:IIR} shows that the imperfect information retrieval module significantly enhances the performance of GPT-4 and GPT-4o, increasing the number of imperfect information games passed under \passATfive. Notably, GPT-4o benefits the most from the module.
In contrast, for GPT-3.5, no performance improvement was observed, suggesting that LLM itself is at fault.
Finally, we note that for GPT-4 and GPT-4o, translation of perfect information games was also better with the module. 
We attribute this to the reasoning process (outlined in Appendix~\ref{app: stage 1 output}) generated in stage one, which aids the LLM in stage two by improving its ability to identify game types and the associated tree structures.

\subsection{Experiments with \citet{mensfelt2024autoformalization} Games}
\label{sec:more_games}
We further evaluate the performance of our framework using the 112 game descriptions from \citet{mensfelt2024autoformalization}, as discussed in Section~\ref{sec: dataset}.
For a comparison with their experimental results, we adopt their approach to regeneration attempts: rather than always performing a fixed number of attempts, they stop on the first successful attempt, or give up after 5 failed attempts.
 
\citet{mensfelt2024autoformalization} used a logic programming approach. 
Across the five underlying bimatrix games in their dataset,
they achieved 100\% accuracy for the Hawk-Dove and Stag Hunt games, but encountered errors 
for descriptions of the Prisoner’s Dilemma, Battle of the Sexes, and Matching Pennies games.
With the same setup, our approach correctly translated all 112 of their game descriptions.
For 102 of these 112 cases, the first correct translation was on the first attempt, the remaining 10 were correctly translated on the second attempt.
This demonstrates the robustness of our method to variations in game descriptions in the 
context of 2x2 games (as future work, it would be good to explore this for more complex 
games too).
We note that we were able to immediately apply our approach to their setting because EFGs are more general than bimatrix games.
In contrast, their approach cannot be directly applied to generate EFG files, so we
could not test it on EFG file generation with our custom dataset.


\section{Conclusion and Discussion}
We introduced a two-stage framework for translating game descriptions in natural language into extensive-form representations in game theory, leveraging LLMs, in-context learning, and code generation using the pygambit library.
In the first stage, we address imperfect information by designing a module that identifies information sets and the corresponding partial game tree structure. 
In the second stage, the output from stage one is used, along with a self-debugging module, to generate a complete EFG using pygambit.
We evaluate the framework's overall performance, as well as its individual components, across three LLMs on 18+112=130 game descriptions spanning 21 different games\footnote{Bach or Stravinksy
in the custom data set is the Battle of the Sexes from~\citet{mensfelt2024autoformalization}, and Rock Paper Scissors appears in both datasets, so we have 18-2+5=21 distinct games overall.}.
Our experimental results show that the framework significantly outperforms baseline models in generating accurate EFGs, with each module playing a critical role in its success. 

One potential direction for future work is to move beyond manual consistency checks and develop a robust, automated check.
Specifically, automated validation may require a ``suite'' of checks covering various aspects of consistency. 
For instance, $\alpha$-rank \cite{omidshafiei2019alpha} could be applied to compare rankings of strategies between target
and generated EFGs; more generally different types of strategic equivalence could used to design checks. Other checks could involve game feature extraction, such as identifying the number of players, possible outcomes, and information sets. 
Note that it is essential to design these checks carefully to avoid introducing sources of error.

Another direction of further work relates to the fact that many games are actually parameterized.
For instance, parameters include the number of stages in the Centipede game, the pile sizes in Nim, or both the number of stages and the discount factor in bargaining games. Our framework takes a game description and produces an EFG file that is consistent with that description. 
A natural extension would take instead a description of a parameterized game family, and
would then generate, instead of a single EFG file, a python function, for example, using
pygambit, that takes game parameters as inputs and generates a corresponding EFG
file.
The development of such an extension would benefit from the previously mentioned automated checking of the consistency
of an EFG file (given a parameterized game description and a specific choice of parameters).

Finally, we believe it is important to explore game description translation for larger and more complex games.
In our custom dataset, Kuhn Poker, with 25 decision nodes and 30 terminal nodes, was the largest game.
%
To tackle larger games, two significant challenges must be addressed: the generation method's capacity to handle the increased complexity and our ability to accurately validate the method's outputs. 
Progress on either direction mentioned above could help with this:
solving parametrized games could be part of a divide and conquer approach to translating complex games,
and robust automated checking of translation outputs would help with the second challenge.
Additionally, alternatives to in-context learning such as supervised fine-tuning could also be effective for solving such problem if a suitable game dataset is available.
 


\bibliographystyle{ACM-Reference-Format} 
\bibliography{main}


\begin{thebibliography}{54}


\ifx \showCODEN    \undefined \def \showCODEN     #1{\unskip}     \fi
\ifx \showDOI      \undefined \def \showDOI       #1{#1}\fi
\ifx \showISBNx    \undefined \def \showISBNx     #1{\unskip}     \fi
\ifx \showISBNxiii \undefined \def \showISBNxiii  #1{\unskip}     \fi
\ifx \showISSN     \undefined \def \showISSN      #1{\unskip}     \fi
\ifx \showLCCN     \undefined \def \showLCCN      #1{\unskip}     \fi
\ifx \shownote     \undefined \def \shownote      #1{#1}          \fi
\ifx \showarticletitle \undefined \def \showarticletitle #1{#1}   \fi
\ifx \showURL      \undefined \def \showURL       {\relax}        \fi
\providecommand\bibfield[2]{#2}
\providecommand\bibinfo[2]{#2}
\providecommand\natexlab[1]{#1}
\providecommand\showeprint[2][]{arXiv:#2}

\bibitem[\protect\citeauthoryear{Achiam, Adler, Agarwal, Ahmad, Akkaya, Aleman, Almeida, Altenschmidt, Altman, Anadkat, et~al\mbox{.}}{Achiam et~al\mbox{.}}{2023}]%
        {achiam2023gpt}
\bibfield{author}{\bibinfo{person}{Josh Achiam}, \bibinfo{person}{Steven Adler}, \bibinfo{person}{Sandhini Agarwal}, \bibinfo{person}{Lama Ahmad}, \bibinfo{person}{Ilge Akkaya}, \bibinfo{person}{Florencia~Leoni Aleman}, \bibinfo{person}{Diogo Almeida}, \bibinfo{person}{Janko Altenschmidt}, \bibinfo{person}{Sam Altman}, \bibinfo{person}{Shyamal Anadkat}, {et~al\mbox{.}}} \bibinfo{year}{2023}\natexlab{}.
\newblock \showarticletitle{{GPT}-4 Technical Report}.
\newblock \bibinfo{journal}{\emph{arXiv preprint arXiv:2303.08774}} (\bibinfo{year}{2023}).
\newblock


\bibitem[\protect\citeauthoryear{Akata, Schulz, Coda-Forno, Oh, Bethge, and Schulz}{Akata et~al\mbox{.}}{2023}]%
        {akata2023playing}
\bibfield{author}{\bibinfo{person}{Elif Akata}, \bibinfo{person}{Lion Schulz}, \bibinfo{person}{Julian Coda-Forno}, \bibinfo{person}{Seong~Joon Oh}, \bibinfo{person}{Matthias Bethge}, {and} \bibinfo{person}{Eric Schulz}.} \bibinfo{year}{2023}\natexlab{}.
\newblock \showarticletitle{Playing repeated games with Large Language Models}.
\newblock \bibinfo{journal}{\emph{arXiv preprint arXiv:2305.16867}} (\bibinfo{year}{2023}).
\newblock


\bibitem[\protect\citeauthoryear{Aumann, Hart, and Perry}{Aumann et~al\mbox{.}}{1997}]%
        {aumann1997absent}
\bibfield{author}{\bibinfo{person}{Robert~J Aumann}, \bibinfo{person}{Sergiu Hart}, {and} \bibinfo{person}{Motty Perry}.} \bibinfo{year}{1997}\natexlab{}.
\newblock \showarticletitle{The Absent-Minded Driver}.
\newblock \bibinfo{journal}{\emph{Games and Economic Behavior}} \bibinfo{volume}{20}, \bibinfo{number}{1} (\bibinfo{year}{1997}), \bibinfo{pages}{102--116}.
\newblock


\bibitem[\protect\citeauthoryear{Bagwell}{Bagwell}{1995}]%
        {bagwell1995commitment}
\bibfield{author}{\bibinfo{person}{Kyle Bagwell}.} \bibinfo{year}{1995}\natexlab{}.
\newblock \showarticletitle{Commitment and Observability in Games}.
\newblock \bibinfo{journal}{\emph{Games and Economic Behavior}} \bibinfo{volume}{8}, \bibinfo{number}{2} (\bibinfo{year}{1995}), \bibinfo{pages}{271--280}.
\newblock


\bibitem[\protect\citeauthoryear{Brown}{Brown}{2020}]%
        {brown2020language}
\bibfield{author}{\bibinfo{person}{Tom~B Brown}.} \bibinfo{year}{2020}\natexlab{}.
\newblock \showarticletitle{Language Models are Few-Shot Learners}.
\newblock \bibinfo{journal}{\emph{arXiv preprint arXiv:2005.14165}} (\bibinfo{year}{2020}).
\newblock


\bibitem[\protect\citeauthoryear{Chen, Tworek, Jun, Yuan, Pinto, Kaplan, Edwards, Burda, Joseph, Brockman, et~al\mbox{.}}{Chen et~al\mbox{.}}{2021}]%
        {chen2021evaluating}
\bibfield{author}{\bibinfo{person}{Mark Chen}, \bibinfo{person}{Jerry Tworek}, \bibinfo{person}{Heewoo Jun}, \bibinfo{person}{Qiming Yuan}, \bibinfo{person}{Henrique Ponde De~Oliveira Pinto}, \bibinfo{person}{Jared Kaplan}, \bibinfo{person}{Harri Edwards}, \bibinfo{person}{Yuri Burda}, \bibinfo{person}{Nicholas Joseph}, \bibinfo{person}{Greg Brockman}, {et~al\mbox{.}}} \bibinfo{year}{2021}\natexlab{}.
\newblock \showarticletitle{Evaluating Large Language Models Trained on Code}.
\newblock \bibinfo{journal}{\emph{arXiv preprint arXiv:2107.03374}} (\bibinfo{year}{2021}).
\newblock


\bibitem[\protect\citeauthoryear{Christopoulou, Lampouras, Gritta, Zhang, Guo, Li, Zhang, Xiao, Shen, Li, et~al\mbox{.}}{Christopoulou et~al\mbox{.}}{2022}]%
        {fenia2022pangu}
\bibfield{author}{\bibinfo{person}{Fenia Christopoulou}, \bibinfo{person}{Gerasimos Lampouras}, \bibinfo{person}{Milan Gritta}, \bibinfo{person}{Guchun Zhang}, \bibinfo{person}{Yinpeng Guo}, \bibinfo{person}{Zhongqi Li}, \bibinfo{person}{Qi Zhang}, \bibinfo{person}{Meng Xiao}, \bibinfo{person}{Bo Shen}, \bibinfo{person}{Lin Li}, {et~al\mbox{.}}} \bibinfo{year}{2022}\natexlab{}.
\newblock \showarticletitle{PanGu-Coder: Program Synthesis with Function-Level Language Modeling}.
\newblock \bibinfo{journal}{\emph{arXiv preprint arXiv:2207.11280}} (\bibinfo{year}{2022}).
\newblock


\bibitem[\protect\citeauthoryear{Crowley and Siegler}{Crowley and Siegler}{1993}]%
        {crowley1993flexible}
\bibfield{author}{\bibinfo{person}{Kevin Crowley} {and} \bibinfo{person}{Robert~S Siegler}.} \bibinfo{year}{1993}\natexlab{}.
\newblock \showarticletitle{Flexible Strategy Use in Young Children's Tic‐Tac‐Toe}.
\newblock \bibinfo{journal}{\emph{Cognitive Science}} \bibinfo{volume}{17}, \bibinfo{number}{4} (\bibinfo{year}{1993}), \bibinfo{pages}{531--561}.
\newblock


\bibitem[\protect\citeauthoryear{Daskalakis, Gemp, Jiang, Leme, Papadimitriou, and Piliouras}{Daskalakis et~al\mbox{.}}{2024}]%
        {daskalakis2024charting}
\bibfield{author}{\bibinfo{person}{Constantinos Daskalakis}, \bibinfo{person}{Ian Gemp}, \bibinfo{person}{Yanchen Jiang}, \bibinfo{person}{Renato~Paes Leme}, \bibinfo{person}{Christos Papadimitriou}, {and} \bibinfo{person}{Georgios Piliouras}.} \bibinfo{year}{2024}\natexlab{}.
\newblock \showarticletitle{Charting the Shapes of Stories with Game Theory}. In \bibinfo{booktitle}{\emph{Proceedings of Advances in Neural Information Processing Systems (NeurIPS)}}.
\newblock


\bibitem[\protect\citeauthoryear{Deng, Wang, and Savani}{Deng et~al\mbox{.}}{2025}]%
        {DWS25}
\bibfield{author}{\bibinfo{person}{Shilong Deng}, \bibinfo{person}{Yongzhao Wang}, {and} \bibinfo{person}{Rahul Savani}.} \bibinfo{year}{2025}\natexlab{}.
\newblock \showarticletitle{From Natural Language to Extensive-Form Game Representations}. In \bibinfo{booktitle}{\emph{Proceedings of the International Conference on Autonomous Agents and Multiagent Systems Advances (AAMAS)}}.
\newblock
\newblock
\shownote{To appear.}


\bibitem[\protect\citeauthoryear{Dong, Li, Dai, Zheng, Wu, Chang, Sun, Xu, and Sui}{Dong et~al\mbox{.}}{2022}]%
        {dong2022survey}
\bibfield{author}{\bibinfo{person}{Qingxiu Dong}, \bibinfo{person}{Lei Li}, \bibinfo{person}{Damai Dai}, \bibinfo{person}{Ce Zheng}, \bibinfo{person}{Zhiyong Wu}, \bibinfo{person}{Baobao Chang}, \bibinfo{person}{Xu Sun}, \bibinfo{person}{Jingjing Xu}, {and} \bibinfo{person}{Zhifang Sui}.} \bibinfo{year}{2022}\natexlab{}.
\newblock \showarticletitle{A Survey on In-context Learning}.
\newblock \bibinfo{journal}{\emph{arXiv preprint arXiv:2301.00234}} (\bibinfo{year}{2022}).
\newblock


\bibitem[\protect\citeauthoryear{Dubey, Jauhri, Pandey, Kadian, Al-Dahle, Letman, Mathur, Schelten, Yang, Fan, et~al\mbox{.}}{Dubey et~al\mbox{.}}{2024}]%
        {dubey2024llama}
\bibfield{author}{\bibinfo{person}{Abhimanyu Dubey}, \bibinfo{person}{Abhinav Jauhri}, \bibinfo{person}{Abhinav Pandey}, \bibinfo{person}{Abhishek Kadian}, \bibinfo{person}{Ahmad Al-Dahle}, \bibinfo{person}{Aiesha Letman}, \bibinfo{person}{Akhil Mathur}, \bibinfo{person}{Alan Schelten}, \bibinfo{person}{Amy Yang}, \bibinfo{person}{Angela Fan}, {et~al\mbox{.}}} \bibinfo{year}{2024}\natexlab{}.
\newblock \showarticletitle{The Llama 3 Herd of Models}.
\newblock \bibinfo{journal}{\emph{arXiv preprint arXiv:2407.21783}} (\bibinfo{year}{2024}).
\newblock


\bibitem[\protect\citeauthoryear{Fan, Chen, Jin, and He}{Fan et~al\mbox{.}}{2024}]%
        {fan2024can}
\bibfield{author}{\bibinfo{person}{Caoyun Fan}, \bibinfo{person}{Jindou Chen}, \bibinfo{person}{Yaohui Jin}, {and} \bibinfo{person}{Hao He}.} \bibinfo{year}{2024}\natexlab{}.
\newblock \showarticletitle{Can Large Language Models Serve as Rational Players in Game Theory? {A} Systematic Analysis}. In \bibinfo{booktitle}{\emph{Proceedings of the AAAI Conference on Artificial Intelligence (AAAI)}}. \bibinfo{pages}{17960--17967}.
\newblock


\bibitem[\protect\citeauthoryear{Fu, Peng, Khot, and Lapata}{Fu et~al\mbox{.}}{2023}]%
        {fu2023improving}
\bibfield{author}{\bibinfo{person}{Yao Fu}, \bibinfo{person}{Hao Peng}, \bibinfo{person}{Tushar Khot}, {and} \bibinfo{person}{Mirella Lapata}.} \bibinfo{year}{2023}\natexlab{}.
\newblock \showarticletitle{Improving Language Model Negotiation with Self-Play and In-Context Learning from AI Feedback}.
\newblock \bibinfo{journal}{\emph{arXiv preprint arXiv:2305.10142}} (\bibinfo{year}{2023}).
\newblock


\bibitem[\protect\citeauthoryear{Gemp, Patel, Bachrach, Lanctot, Dasagi, Marris, Piliouras, Liu, and Tuyls}{Gemp et~al\mbox{.}}{2024}]%
        {gemp2024states}
\bibfield{author}{\bibinfo{person}{Ian Gemp}, \bibinfo{person}{Roma Patel}, \bibinfo{person}{Yoram Bachrach}, \bibinfo{person}{Marc Lanctot}, \bibinfo{person}{Vibhavari Dasagi}, \bibinfo{person}{Luke Marris}, \bibinfo{person}{Georgios Piliouras}, \bibinfo{person}{Siqi Liu}, {and} \bibinfo{person}{Karl Tuyls}.} \bibinfo{year}{2024}\natexlab{}.
\newblock \showarticletitle{Steering Language Models with Game-Theoretic Solvers}. In \bibinfo{booktitle}{\emph{Agentic Markets Workshop at International Conference on Machine Learning (AMW@ICML)}}.
\newblock


\bibitem[\protect\citeauthoryear{Gibbons}{Gibbons}{1992}]%
        {gibbons1992game}
\bibfield{author}{\bibinfo{person}{Robert Gibbons}.} \bibinfo{year}{1992}\natexlab{}.
\newblock \bibinfo{booktitle}{\emph{Game Theory for Applied Economists}}.
\newblock \bibinfo{publisher}{Princeton University Press}.
\newblock


\bibitem[\protect\citeauthoryear{Hahn, Schmitt, Tillman, Metzger, Siber, and Finkbeiner}{Hahn et~al\mbox{.}}{2022}]%
        {hahn2022formal}
\bibfield{author}{\bibinfo{person}{Christopher Hahn}, \bibinfo{person}{Frederik Schmitt}, \bibinfo{person}{Julia~J Tillman}, \bibinfo{person}{Niklas Metzger}, \bibinfo{person}{Julian Siber}, {and} \bibinfo{person}{Bernd Finkbeiner}.} \bibinfo{year}{2022}\natexlab{}.
\newblock \showarticletitle{Formal Specifications from Natural Language}.
\newblock \bibinfo{journal}{\emph{arXiv preprint arXiv:2206.01962}} (\bibinfo{year}{2022}).
\newblock


\bibitem[\protect\citeauthoryear{Holtzman, Buys, Du, Forbes, and Choi}{Holtzman et~al\mbox{.}}{2020}]%
        {HoltzmanBDFC20}
\bibfield{author}{\bibinfo{person}{Ari Holtzman}, \bibinfo{person}{Jan Buys}, \bibinfo{person}{Li Du}, \bibinfo{person}{Maxwell Forbes}, {and} \bibinfo{person}{Yejin Choi}.} \bibinfo{year}{2020}\natexlab{}.
\newblock \showarticletitle{The Curious Case of Neural Text Degeneration}. In \bibinfo{booktitle}{\emph{Proceedings of the International Conference on Learning Representations {ICLR}}}.
\newblock


\bibitem[\protect\citeauthoryear{Huang, Wei, Yue, Zhao, Chen, Li, Jiang, Chang, Zhang, Zhang, et~al\mbox{.}}{Huang et~al\mbox{.}}{2023}]%
        {huang2023dsqa}
\bibfield{author}{\bibinfo{person}{Dengrong Huang}, \bibinfo{person}{Zizhong Wei}, \bibinfo{person}{Aizhen Yue}, \bibinfo{person}{Xuan Zhao}, \bibinfo{person}{Zhaoliang Chen}, \bibinfo{person}{Rui Li}, \bibinfo{person}{Kai Jiang}, \bibinfo{person}{Bingxin Chang}, \bibinfo{person}{Qilai Zhang}, \bibinfo{person}{Sijia Zhang}, {et~al\mbox{.}}} \bibinfo{year}{2023}\natexlab{}.
\newblock \showarticletitle{DSQA-LLM: Domain-Specific Intelligent Question Answering Based on Large Language Model}. In \bibinfo{booktitle}{\emph{Proceedings of International Conference on AI-generated Content (AIGC)}}. \bibinfo{pages}{170--180}.
\newblock


\bibitem[\protect\citeauthoryear{Jatova, Smith, and Wilson}{Jatova et~al\mbox{.}}{2024}]%
        {jatova2024employing}
\bibfield{author}{\bibinfo{person}{Lucas Jatova}, \bibinfo{person}{Jacob Smith}, {and} \bibinfo{person}{Alexander Wilson}.} \bibinfo{year}{2024}\natexlab{}.
\newblock \showarticletitle{Employing Game Theory for Mitigating Adversarial-Induced Content Toxicity in Generative Large Language Models}.
\newblock \bibinfo{journal}{\emph{TechRxiv}} (\bibinfo{year}{2024}).
\newblock


\bibitem[\protect\citeauthoryear{Kuhn}{Kuhn}{1950}]%
        {kuhn1950simplified}
\bibfield{author}{\bibinfo{person}{Harold~W Kuhn}.} \bibinfo{year}{1950}\natexlab{}.
\newblock \showarticletitle{A simplified two-person poker}.
\newblock \bibinfo{journal}{\emph{Contributions to the Theory of Games}} \bibinfo{volume}{1}, \bibinfo{number}{97-103} (\bibinfo{year}{1950}), \bibinfo{pages}{2}.
\newblock


\bibitem[\protect\citeauthoryear{Le, Wang, Gotmare, Savarese, and Hoi}{Le et~al\mbox{.}}{2022}]%
        {le2022coderl}
\bibfield{author}{\bibinfo{person}{Hung Le}, \bibinfo{person}{Yue Wang}, \bibinfo{person}{Akhilesh~Deepak Gotmare}, \bibinfo{person}{Silvio Savarese}, {and} \bibinfo{person}{Steven Chu~Hong Hoi}.} \bibinfo{year}{2022}\natexlab{}.
\newblock \showarticletitle{CodeRL: Mastering Code Generation through Pretrained Models and Deep Reinforcement Learning}. In \bibinfo{booktitle}{\emph{Proceedings of Advances in Neural Information Processing Systems (NeurIPS)}}, Vol.~\bibinfo{volume}{35}. \bibinfo{pages}{21314--21328}.
\newblock


\bibitem[\protect\citeauthoryear{Leite, Arruda, Antonino, Sampaio, and Roscoe}{Leite et~al\mbox{.}}{2024}]%
        {leite2024extracting}
\bibfield{author}{\bibinfo{person}{Gabriel Leite}, \bibinfo{person}{Filipe Arruda}, \bibinfo{person}{Pedro Antonino}, \bibinfo{person}{Augusto Sampaio}, {and} \bibinfo{person}{AW Roscoe}.} \bibinfo{year}{2024}\natexlab{}.
\newblock \showarticletitle{Extracting Formal Smart-Contract Specifications from Natural Language with {LLM}s}. In \bibinfo{booktitle}{\emph{Proceedings of the International Conference on Formal Aspects of Component Software (FACS)}}, Vol.~\bibinfo{volume}{15189}. \bibinfo{pages}{109--126}.
\newblock


\bibitem[\protect\citeauthoryear{Leong and Barbosa}{Leong and Barbosa}{2023}]%
        {leong2023translating}
\bibfield{author}{\bibinfo{person}{Iat~Tou Leong} {and} \bibinfo{person}{Raul Barbosa}.} \bibinfo{year}{2023}\natexlab{}.
\newblock \showarticletitle{Translating Natural Language Requirements to Formal Specifications: A Study on GPT and Symbolic NLP}. In \bibinfo{booktitle}{\emph{Proceedings of the IEEE/IFIP International Conference on Dependable Systems and Networks Workshops (DSN-W)}}. \bibinfo{pages}{259--262}.
\newblock


\bibitem[\protect\citeauthoryear{Levin}{Levin}{2002}]%
        {levin2002extensive}
\bibfield{author}{\bibinfo{person}{Jonathan Levin}.} \bibinfo{year}{2002}\natexlab{}.
\newblock \bibinfo{title}{Extensive Form Games}.
\newblock
\newblock
\urldef\tempurl%
\url{https://web.stanford.edu/~jdlevin/Econ%20203/ExtensiveForm.pdf}
\showURL{%
\tempurl}
\newblock
\shownote{Accessed: 2024-11-03.}


\bibitem[\protect\citeauthoryear{Li, Choi, Chung, Kushman, Schrittwieser, Leblond, Eccles, Keeling, Gimeno, Dal~Lago, et~al\mbox{.}}{Li et~al\mbox{.}}{2022}]%
        {li2022competition}
\bibfield{author}{\bibinfo{person}{Yujia Li}, \bibinfo{person}{David Choi}, \bibinfo{person}{Junyoung Chung}, \bibinfo{person}{Nate Kushman}, \bibinfo{person}{Julian Schrittwieser}, \bibinfo{person}{R{\'e}mi Leblond}, \bibinfo{person}{Tom Eccles}, \bibinfo{person}{James Keeling}, \bibinfo{person}{Felix Gimeno}, \bibinfo{person}{Agustin Dal~Lago}, {et~al\mbox{.}}} \bibinfo{year}{2022}\natexlab{}.
\newblock \showarticletitle{Competition-Level Code Generation with AlphaCode}.
\newblock \bibinfo{journal}{\emph{Science}} \bibinfo{volume}{378}, \bibinfo{number}{6624} (\bibinfo{year}{2022}), \bibinfo{pages}{1092--1097}.
\newblock


\bibitem[\protect\citeauthoryear{Li, Fan, Gu, Li, Duan, Dong, Liu, and Wang}{Li et~al\mbox{.}}{2024}]%
        {li2024flexkbqa}
\bibfield{author}{\bibinfo{person}{Zhenyu Li}, \bibinfo{person}{Sunqi Fan}, \bibinfo{person}{Yu Gu}, \bibinfo{person}{Xiuxing Li}, \bibinfo{person}{Zhichao Duan}, \bibinfo{person}{Bowen Dong}, \bibinfo{person}{Ning Liu}, {and} \bibinfo{person}{Jianyong Wang}.} \bibinfo{year}{2024}\natexlab{}.
\newblock \showarticletitle{FlexKBQA: {A} Flexible LLM-Powered Framework for Few-Shot Knowledge Base Question Answering}. In \bibinfo{booktitle}{\emph{Proceedings of the AAAI Conference on Artificial Intelligence (AAAI)}}. \bibinfo{pages}{18608--18616}.
\newblock


\bibitem[\protect\citeauthoryear{Liu, Zhu, Xiao, FU, Han, Wei, and Ye}{Liu et~al\mbox{.}}{2023}]%
        {liu2023rltf}
\bibfield{author}{\bibinfo{person}{Jiate Liu}, \bibinfo{person}{Yiqin Zhu}, \bibinfo{person}{Kaiwen Xiao}, \bibinfo{person}{QIANG FU}, \bibinfo{person}{Xiao Han}, \bibinfo{person}{Yang Wei}, {and} \bibinfo{person}{Deheng Ye}.} \bibinfo{year}{2023}\natexlab{}.
\newblock \showarticletitle{{RLTF}: Reinforcement Learning from Unit Test Feedback}.
\newblock \bibinfo{journal}{\emph{Transactions on Machine Learning Research}} (\bibinfo{year}{2023}).
\newblock
\showISSN{2835-8856}


\bibitem[\protect\citeauthoryear{Lor{\`e} and Heydari}{Lor{\`e} and Heydari}{2023}]%
        {lore2023strategic}
\bibfield{author}{\bibinfo{person}{Nunzio Lor{\`e}} {and} \bibinfo{person}{Babak Heydari}.} \bibinfo{year}{2023}\natexlab{}.
\newblock \showarticletitle{Strategic Behavior of Large Language Models: Game Structure vs. Contextual Framing}.
\newblock \bibinfo{journal}{\emph{arXiv preprint arXiv:2309.05898}} (\bibinfo{year}{2023}).
\newblock


\bibitem[\protect\citeauthoryear{Ma, Yang, Gao, Ci, Gao, Pan, and Yang}{Ma et~al\mbox{.}}{2023}]%
        {ma2023red}
\bibfield{author}{\bibinfo{person}{Chengdong Ma}, \bibinfo{person}{Ziran Yang}, \bibinfo{person}{Minquan Gao}, \bibinfo{person}{Hai Ci}, \bibinfo{person}{Jun Gao}, \bibinfo{person}{Xuehai Pan}, {and} \bibinfo{person}{Yaodong Yang}.} \bibinfo{year}{2023}\natexlab{}.
\newblock \showarticletitle{Red Teaming Game: A Game-Theoretic Framework for Red Teaming Language Models}.
\newblock \bibinfo{journal}{\emph{arXiv preprint arXiv:2310.00322}} (\bibinfo{year}{2023}).
\newblock


\bibitem[\protect\citeauthoryear{Manea}{Manea}{2016}]%
        {Manea2016}
\bibfield{author}{\bibinfo{person}{Mihai Manea}.} \bibinfo{year}{2016}\natexlab{}.
\newblock \bibinfo{title}{MIT Extensive Form Games Lecture Slides}.
\newblock \bibinfo{howpublished}{\url{https://ocw.mit.edu/courses/14-16-strategy-and-information-spring-2016/8aba65347a9edb17b1e718c3c221e42a_MIT14_16S16_Form_Games.pdf}}.
\newblock
\newblock
\shownote{Accessed: 2024-10-16.}


\bibitem[\protect\citeauthoryear{Mao, Cai, Xia, Wu, Wang, Wang, Ge, and Wei}{Mao et~al\mbox{.}}{2023}]%
        {mao2023alympics}
\bibfield{author}{\bibinfo{person}{Shaoguang Mao}, \bibinfo{person}{Yuzhe Cai}, \bibinfo{person}{Yan Xia}, \bibinfo{person}{Wenshan Wu}, \bibinfo{person}{Xun Wang}, \bibinfo{person}{Fengyi Wang}, \bibinfo{person}{Tao Ge}, {and} \bibinfo{person}{Furu Wei}.} \bibinfo{year}{2023}\natexlab{}.
\newblock \showarticletitle{ALYMPICS: Language Agents Meet Game Theory}.
\newblock \bibinfo{journal}{\emph{arXiv preprint arXiv:2311.03220}} (\bibinfo{year}{2023}).
\newblock


\bibitem[\protect\citeauthoryear{Maschler, Zamir, and Solan}{Maschler et~al\mbox{.}}{2020}]%
        {maschler2020game}
\bibfield{author}{\bibinfo{person}{Michael Maschler}, \bibinfo{person}{Shmuel Zamir}, {and} \bibinfo{person}{Eilon Solan}.} \bibinfo{year}{2020}\natexlab{}.
\newblock \bibinfo{booktitle}{\emph{Game Theory}}.
\newblock \bibinfo{publisher}{Cambridge University Press}.
\newblock


\bibitem[\protect\citeauthoryear{Mensfelt, Stathis, and Trencsenyi}{Mensfelt et~al\mbox{.}}{2024a}]%
        {mensfelt2024autoformalization}
\bibfield{author}{\bibinfo{person}{Agnieszka Mensfelt}, \bibinfo{person}{Kostas Stathis}, {and} \bibinfo{person}{Vince Trencsenyi}.} \bibinfo{year}{2024}\natexlab{a}.
\newblock \showarticletitle{Autoformalization of Game Descriptions using Large Language Models}.
\newblock \bibinfo{journal}{\emph{arXiv preprint arXiv:2409.12300}} (\bibinfo{year}{2024}).
\newblock


\bibitem[\protect\citeauthoryear{Mensfelt, Stathis, and Trencsenyi}{Mensfelt et~al\mbox{.}}{2024b}]%
        {mensfelt2024autoformalizing}
\bibfield{author}{\bibinfo{person}{Agnieszka Mensfelt}, \bibinfo{person}{Kostas Stathis}, {and} \bibinfo{person}{Vince Trencsenyi}.} \bibinfo{year}{2024}\natexlab{b}.
\newblock \showarticletitle{Autoformalizing and Simulating Game-Theoretic Scenarios using LLM-augmented Agents}.
\newblock \bibinfo{journal}{\emph{arXiv preprint arXiv:2412.08805}} (\bibinfo{year}{2024}).
\newblock


\bibitem[\protect\citeauthoryear{Nash}{Nash}{1950}]%
        {nash1950bargaining}
\bibfield{author}{\bibinfo{person}{John~F Nash}.} \bibinfo{year}{1950}\natexlab{}.
\newblock \showarticletitle{The Bargaining Problem}.
\newblock \bibinfo{journal}{\emph{Econometrica}} \bibinfo{volume}{18}, \bibinfo{number}{2} (\bibinfo{year}{1950}), \bibinfo{pages}{155--162}.
\newblock


\bibitem[\protect\citeauthoryear{Omidshafiei, Papadimitriou, Piliouras, Tuyls, Rowland, Lespiau, Czarnecki, Lanctot, Perolat, and Munos}{Omidshafiei et~al\mbox{.}}{2019}]%
        {omidshafiei2019alpha}
\bibfield{author}{\bibinfo{person}{Shayegan Omidshafiei}, \bibinfo{person}{Christos Papadimitriou}, \bibinfo{person}{Georgios Piliouras}, \bibinfo{person}{Karl Tuyls}, \bibinfo{person}{Mark Rowland}, \bibinfo{person}{Jean-Baptiste Lespiau}, \bibinfo{person}{Wojciech~M Czarnecki}, \bibinfo{person}{Marc Lanctot}, \bibinfo{person}{Julien Perolat}, {and} \bibinfo{person}{Remi Munos}.} \bibinfo{year}{2019}\natexlab{}.
\newblock \showarticletitle{$\alpha$-rank: Multi-agent evaluation by evolution}.
\newblock \bibinfo{journal}{\emph{Scientific Reports}}  \bibinfo{volume}{9, 9937} (\bibinfo{year}{2019}), \bibinfo{pages}{29}.
\newblock


\bibitem[\protect\citeauthoryear{OpenAI}{OpenAI}{2024}]%
        {openai2024gpt4o}
\bibfield{author}{\bibinfo{person}{OpenAI}.} \bibinfo{year}{2024}\natexlab{}.
\newblock \bibinfo{title}{GPT-4o System Card}.
\newblock
\newblock
\urldef\tempurl%
\url{https://cdn.openai.com/gpt-4o-system-card.pdf}
\showURL{%
\tempurl}
\newblock
\shownote{Accessed: 2024-10-13.}


\bibitem[\protect\citeauthoryear{Osborne}{Osborne}{1994}]%
        {osborne1994course}
\bibfield{author}{\bibinfo{person}{Martin~J Osborne}.} \bibinfo{year}{1994}\natexlab{}.
\newblock \bibinfo{booktitle}{\emph{A Course in Game Theory}}.
\newblock \bibinfo{publisher}{MIT Press}.
\newblock


\bibitem[\protect\citeauthoryear{Poje, Brcic, Kovac, and Babac}{Poje et~al\mbox{.}}{2024}]%
        {poje2024effect}
\bibfield{author}{\bibinfo{person}{Kristijan Poje}, \bibinfo{person}{Mario Brcic}, \bibinfo{person}{Mihael Kovac}, {and} \bibinfo{person}{Marina~Bagic Babac}.} \bibinfo{year}{2024}\natexlab{}.
\newblock \showarticletitle{Effect of Private Deliberation: Deception of Large Language Models in Game Play}.
\newblock \bibinfo{journal}{\emph{Entropy}} \bibinfo{volume}{26}, \bibinfo{number}{6} (\bibinfo{year}{2024}), \bibinfo{pages}{524}.
\newblock


\bibitem[\protect\citeauthoryear{Rosenthal}{Rosenthal}{1981}]%
        {rosenthal1981games}
\bibfield{author}{\bibinfo{person}{Robert~W Rosenthal}.} \bibinfo{year}{1981}\natexlab{}.
\newblock \showarticletitle{Games of Perfect Information, Predatory Pricing and the Chain-Store Paradox}.
\newblock \bibinfo{journal}{\emph{Journal of Economic Theory}} \bibinfo{volume}{25}, \bibinfo{number}{1} (\bibinfo{year}{1981}), \bibinfo{pages}{92--100}.
\newblock


\bibitem[\protect\citeauthoryear{Roziere, Gehring, Gloeckle, Sootla, Gat, Tan, Adi, Liu, Sauvestre, Remez, et~al\mbox{.}}{Roziere et~al\mbox{.}}{2023}]%
        {roziere2023code}
\bibfield{author}{\bibinfo{person}{Baptiste Roziere}, \bibinfo{person}{Jonas Gehring}, \bibinfo{person}{Fabian Gloeckle}, \bibinfo{person}{Sten Sootla}, \bibinfo{person}{Itai Gat}, \bibinfo{person}{Xiaoqing~Ellen Tan}, \bibinfo{person}{Yossi Adi}, \bibinfo{person}{Jingyu Liu}, \bibinfo{person}{Romain Sauvestre}, \bibinfo{person}{Tal Remez}, {et~al\mbox{.}}} \bibinfo{year}{2023}\natexlab{}.
\newblock \showarticletitle{Code {L}lama: Open Foundation Models for Code}.
\newblock \bibinfo{journal}{\emph{arXiv preprint arXiv:2308.12950}} (\bibinfo{year}{2023}).
\newblock


\bibitem[\protect\citeauthoryear{Savani and Turocy}{Savani and Turocy}{2024}]%
        {savani2024gambit}
\bibfield{author}{\bibinfo{person}{Rahul Savani} {and} \bibinfo{person}{Theodore~L. Turocy}.} \bibinfo{year}{2024}\natexlab{}.
\newblock \bibinfo{booktitle}{\emph{Gambit: The package for computation in game theory, Version 16.2.0}}.
\newblock
\urldef\tempurl%
\url{http://www.gambit-project.org}
\showURL{%
\tempurl}
\newblock
\shownote{Version 16.2.0.}


\bibitem[\protect\citeauthoryear{Shi, Fang, Zheng, Deng, Chen, and Du}{Shi et~al\mbox{.}}{2023}]%
        {shi2023cooperation}
\bibfield{author}{\bibinfo{person}{Zijing Shi}, \bibinfo{person}{Meng Fang}, \bibinfo{person}{Shunfeng Zheng}, \bibinfo{person}{Shilong Deng}, \bibinfo{person}{Ling Chen}, {and} \bibinfo{person}{Yali Du}.} \bibinfo{year}{2023}\natexlab{}.
\newblock \showarticletitle{Cooperation on the Fly: Exploring Language Agents for Ad Hoc Teamwork in the Avalon Game}.
\newblock \bibinfo{journal}{\emph{arXiv preprint arXiv:2312.17515}} (\bibinfo{year}{2023}).
\newblock


\bibitem[\protect\citeauthoryear{Shinn, Cassano, Gopinath, Narasimhan, and Yao}{Shinn et~al\mbox{.}}{2024}]%
        {shinn2024reflexion}
\bibfield{author}{\bibinfo{person}{Noah Shinn}, \bibinfo{person}{Federico Cassano}, \bibinfo{person}{Ashwin Gopinath}, \bibinfo{person}{Karthik Narasimhan}, {and} \bibinfo{person}{Shunyu Yao}.} \bibinfo{year}{2024}\natexlab{}.
\newblock \showarticletitle{Reflexion: Language Agents with Verbal Reinforcement Learning}. In \bibinfo{booktitle}{\emph{Proceedings of Advances in Neural Information Processing Systems (NeurIPS)}}. \bibinfo{pages}{8634--8652}.
\newblock


\bibitem[\protect\citeauthoryear{Shojaee, Jain, Tipirneni, and Reddy}{Shojaee et~al\mbox{.}}{2023}]%
        {shojaee2023execution}
\bibfield{author}{\bibinfo{person}{Parshin Shojaee}, \bibinfo{person}{Aneesh Jain}, \bibinfo{person}{Sindhu Tipirneni}, {and} \bibinfo{person}{Chandan~K. Reddy}.} \bibinfo{year}{2023}\natexlab{}.
\newblock \showarticletitle{Execution-based Code Generation using Deep Reinforcement Learning}.
\newblock \bibinfo{journal}{\emph{Transactions on Machine Learning Research}} (\bibinfo{year}{2023}).
\newblock
\showISSN{2835-8856}


\bibitem[\protect\citeauthoryear{Silva}{Silva}{2024}]%
        {silva2024large}
\bibfield{author}{\bibinfo{person}{Alonso Silva}.} \bibinfo{year}{2024}\natexlab{}.
\newblock \showarticletitle{Large Language Models Playing Mixed Strategy Nash Equilibrium Games}.
\newblock \bibinfo{journal}{\emph{arXiv preprint arXiv:2406.10574}} (\bibinfo{year}{2024}).
\newblock


\bibitem[\protect\citeauthoryear{Sinervo and Lively}{Sinervo and Lively}{1996}]%
        {sinervo1996rock}
\bibfield{author}{\bibinfo{person}{Barry Sinervo} {and} \bibinfo{person}{Curt~M Lively}.} \bibinfo{year}{1996}\natexlab{}.
\newblock \showarticletitle{The Rock-Paper-Scissors Game and the Evolution of Alternative Male Strategies}.
\newblock \bibinfo{journal}{\emph{Nature}} \bibinfo{volume}{380}, \bibinfo{number}{6571} (\bibinfo{year}{1996}), \bibinfo{pages}{240--243}.
\newblock


\bibitem[\protect\citeauthoryear{Spencer}{Spencer}{1976}]%
        {spencer1974market}
\bibfield{author}{\bibinfo{person}{A.~Michael Spencer}.} \bibinfo{year}{1976}\natexlab{}.
\newblock \showarticletitle{Market Signaling: Informational Transfer in Hiring and Related Screening Processes}.
\newblock \bibinfo{journal}{\emph{Journal of Political Economy}}  \bibinfo{volume}{84} (\bibinfo{year}{1976}), \bibinfo{pages}{200–201}.
\newblock


\bibitem[\protect\citeauthoryear{Tan, Min, Li, Li, Hu, Chen, and Qi}{Tan et~al\mbox{.}}{2023}]%
        {tan2023can}
\bibfield{author}{\bibinfo{person}{Yiming Tan}, \bibinfo{person}{Dehai Min}, \bibinfo{person}{Yu Li}, \bibinfo{person}{Wenbo Li}, \bibinfo{person}{Nan Hu}, \bibinfo{person}{Yongrui Chen}, {and} \bibinfo{person}{Guilin Qi}.} \bibinfo{year}{2023}\natexlab{}.
\newblock \showarticletitle{Can ChatGPT Replace Traditional {KBQA} Models? An In-Depth Analysis of the Question Answering Performance of the {GPT} {LLM} Family}. In \bibinfo{booktitle}{\emph{Proceedings of the International Semantic Web Conference (ISWC)}}, Vol.~\bibinfo{volume}{14265}. \bibinfo{pages}{348--367}.
\newblock


\bibitem[\protect\citeauthoryear{von Stengel}{von Stengel}{2021}]%
        {von2021game}
\bibfield{author}{\bibinfo{person}{Bernhard von Stengel}.} \bibinfo{year}{2021}\natexlab{}.
\newblock \bibinfo{booktitle}{\emph{Game Theory Basics}}.
\newblock \bibinfo{publisher}{Cambridge University Press}.
\newblock


\bibitem[\protect\citeauthoryear{Wei, Wang, Schuurmans, Bosma, brian ichter, Xia, Chi, Le, and Zhou}{Wei et~al\mbox{.}}{2022}]%
        {wei2022chain}
\bibfield{author}{\bibinfo{person}{Jason Wei}, \bibinfo{person}{Xuezhi Wang}, \bibinfo{person}{Dale Schuurmans}, \bibinfo{person}{Maarten Bosma}, \bibinfo{person}{brian ichter}, \bibinfo{person}{Fei Xia}, \bibinfo{person}{Ed~H. Chi}, \bibinfo{person}{Quoc~V Le}, {and} \bibinfo{person}{Denny Zhou}.} \bibinfo{year}{2022}\natexlab{}.
\newblock \showarticletitle{Chain of Thought Prompting Elicits Reasoning in Large Language Models}. In \bibinfo{booktitle}{\emph{Proceedings of Advances in Neural Information Processing Systems (NeurIPS)}}, Vol.~\bibinfo{volume}{35}. \bibinfo{pages}{24824--24837}.
\newblock


\bibitem[\protect\citeauthoryear{Xu, Wang, Li, Luo, Wang, Liu, and Liu}{Xu et~al\mbox{.}}{2023}]%
        {xu2023exploring}
\bibfield{author}{\bibinfo{person}{Yuzhuang Xu}, \bibinfo{person}{Shuo Wang}, \bibinfo{person}{Peng Li}, \bibinfo{person}{Fuwen Luo}, \bibinfo{person}{Xiaolong Wang}, \bibinfo{person}{Weidong Liu}, {and} \bibinfo{person}{Yang Liu}.} \bibinfo{year}{2023}\natexlab{}.
\newblock \showarticletitle{Exploring Large Language Models for Communication Games: An Empirical Study on Werewolf}.
\newblock \bibinfo{journal}{\emph{arXiv preprint arXiv:2309.04658}} (\bibinfo{year}{2023}).
\newblock


\bibitem[\protect\citeauthoryear{Zhai, Shi, Pan, Zhou, Liu, Fang, Ma, Tan, and Zhang}{Zhai et~al\mbox{.}}{2020}]%
        {zhai2020c2s}
\bibfield{author}{\bibinfo{person}{Juan Zhai}, \bibinfo{person}{Yu Shi}, \bibinfo{person}{Minxue Pan}, \bibinfo{person}{Guian Zhou}, \bibinfo{person}{Yongxiang Liu}, \bibinfo{person}{Chunrong Fang}, \bibinfo{person}{Shiqing Ma}, \bibinfo{person}{Lin Tan}, {and} \bibinfo{person}{Xiangyu Zhang}.} \bibinfo{year}{2020}\natexlab{}.
\newblock \showarticletitle{C2S: Translating Natural Language Comments to Formal Program Specifications}. In \bibinfo{booktitle}{\emph{Proceedings of the ACM Joint Meeting on European Software Engineering Conference and Symposium on the Foundations of Software Engineering (FSE)}}. \bibinfo{pages}{25--37}.
\newblock


\end{thebibliography}


\appendix
\section{Game Descriptions}
\label{app:game}
This section presents the descriptions for the nineteen games that are utilized in our evaluation. 


\subsection{A Three-Player Game \cite{maschler2020game}}
This is a game with three players.
In the first stage, a chance event determines either ``A'' or ``B'' with equal probability (1/2 for each). In the second stage, player 1 chooses between "L" and "R" without knowing the outcomcentie of the chance event. If player 1 selects "R," the game ends with a payoff of (1, 1, 1) to all players. In the third stage, if player 1 chooses ``L,'' then player 2 can select either ``l'' or ``r'' without knowing player 1's choice from the previous stage. If player 2 chooses ``r,'' the game ends with a payoff of (2, 2, 2) to all players.
In the fourth stage, if player 2 selects ``l,'' then player 3 can choose between ``a'' and ``b'' without knowing player 2's prior choice. The game concludes with a payoff of (3, 3, 3) once player 3 makes a decision.
Throughout the game, none of the players have knowledge of the previous moves of other players or the chance event.

\subsection{An Imperfect Recall Game~\cite{maschler2020game}}

Consider a three-stage game. In the first stage, a chance event randomly selects either ``L'' or ``R'', each with a probability of 1/2. In the second stage, Player 1 observes this outcome and then selects either l or r. In the third stage, Player 1 must choose between ``A'', and ``B'', but at this point, she has forgotten the outcome of the first stage and only remembers her decision from the second stage. The payoffs in all outcomes are set to 0.

\subsection{Absent-Minded Driver~\cite{aumann1997absent}}

At junction X, an absent-minded driver has two choices: EXIT, reaching destination A with a payoff of 0, or CONTINUE to junction Y. At junction Y, the driver can choose to EXIT, arriving at destination B with a payoff of 4, or CONTINUE to C, which yields a payoff of 1. The key assumption is that the driver cannot tell the difference between junctions X and Y and does not remember if he has already passed one of them.

\subsection{Bach or Stravinsky~\cite{osborne1994course}}

The ``Bach or Stravinsky?'' game involves two players, Alexis and Beverley, who want to attend a concert together. Alexis prefers Bach, while Beverley prefers Stravinsky. Both players make their choice between Bach and Stravinsky simultaneously and independently. If they both choose Bach, Alexis, the Bach enthusiast, receives a higher payoff, while Beverley receives a lower payoff. Conversely, if they both choose Stravinsky, Beverley, who favors Stravinsky, receives a higher payoff, and Alexis receives a lower payoff. If they choose different concerts, neither player receives any payoff.

\subsection{Bagwell~\cite{bagwell1995commitment}}


In this game, Player 1, the leader, first selects a strategy: either ``S'' or ``C.'' Player 2, the follower, then tries to interpret Player 1's choice, but there is a small probability of misinterpretation (e.g., if Player 1 selects S, there’s a 1\% chance that Player 2 will perceive it as C). Based on what Player 2 believes Player 1 chose, Player 2 then picks either ``S'' or ``C.'' The resulting payoffs for each player are based on the chosen combination (Player 1's actual selection, not Player 2's interpretation): if both choose S, the payoff is (A, B); if Player 1 chooses S and Player 2 chooses C, it is (W, D); if Player 1 chooses C and Player 2 chooses S, it is (E, W); and if both choose C, the payoff is (F, F). The payoff rankings from highest to lowest are: E (highest), A, F, W, B, and D (lowest). Assign values to each of these payoffs and set all the outcomes.

\subsection{Centipede~\cite{rosenthal1981games}}

Consider a game with two players, Alice and Bob, where Alice makes the first move. At the start, Alice has two piles of coins in front of her: one pile with 4 coins and another with 1 coin. Each player has two options on their turn: they can either take the larger pile, giving the smaller pile to the other player, or they can push both piles to the other player. Whenever the piles are pushed across the table, the number of coins in each pile doubles. For instance, if Alice chooses to push on her first turn, the piles of 1 and 4 coins are handed over to Bob, increasing to 2 and 8 coins. Bob can then decide either to take the pile of 8 coins and leave 2 for Alice or to push the piles back to Alice, further doubling them to 4 and 16 coins. If neither player takes the coins by the end of the game, Alice will receive the pile with the higher value, and Bob will get the one with the lower value. The game has four moves in total: Alice moves (take or push), Bob moves (take or push, where the final push also doubles the piles before the game ends), Alice moves again, and finally, Bob moves. All actions are visible to both players.

\subsection{Colonial Control~\cite{gibbons1992game}}

There are two players in this scenario: Country A and Country B. The status quo is that Country A receives revenue from Country B. Initially, Country B has the choice to either ``Accept'' this arrangement or ``Rebel.'' If Country B accepts the status quo, Country A then decides whether to ``Tax'' or ``Drop Taxes.'' If Country A continues to tax, it gains 6, while Country B loses 2. If Country A drops the taxes, Country A earns 4, and Country B breaks even with 0. However, if Country B decides to rebel, Country A faces a choice to either ``Grant Independence'' to Country B or attempt to ``Suppress'' the rebellion. Granting independence results in Country A earning 0, while Country B gains 3. Attempting to suppress the rebellion leads to war, with the outcome determined by chance. In the event of war, Country B has a 0.3 probability of winning. If Country B wins the war, it loses 3, while Country A loses 1. If Country B loses the war, it loses 5, and Country A still loses 1.

\subsection{Kuhn Poker~\cite{kuhn1950simplified}}
\label{app:kuhn}
This is a two-player card game between Alice and Bob, using a deck of only three cards: a King, Queen, and Jack. Each player contributes 1 to the pot at the start. 
Each player is dealt one of the three cards, and the third is put aside unseen by a chance node.
There are six possible allocations of the cards, each with probability 1/6: JQ, JK, QJ, QK, KJ, and KQ.
The game proceeds as follows for each of the six possible allocations:\\
Alice can check or bet 1, without knowing the allocation of the chance node.\\
    \hspace*{0.4cm}If Alice checks then Bob can check or bet 1, without knowing \hspace*{0.36cm}Alice’s card.\\
        \hspace*{0.8cm}If Bob checks, the game ends (i.e. the higher card wins 1 \hspace*{0.8cm}from the other player).\\
        \hspace*{0.8cm}If Bob bets then Alice can fold or call, without knowing \hspace*{0.8cm}Bob’s card.\\
            \hspace*{1.2cm}If Alice folds then Bob takes the pot of 3 (i.e. winning \hspace*{1.2cm}1 from Alice).\\
            \hspace*{1.2cm}If Alice calls, the game ends. (i.e. the higher card wins \hspace*{1.1cm} 2 from the other player).\\
    \hspace*{0.4cm}If Alice bets then Bob can fold or call, without knowing Alice’s \hspace*{0.4cm}card.\\
        \hspace*{0.8cm}If Bob folds then Alice takes the pot of 3 (i.e. winning 1 \hspace*{0.8cm}from Bob).\\
        \hspace*{0.8cm}If Bob calls, the game ends. (i.e. the higher card wins 2 \hspace*{0.8cm}from the other player).\\

\subsection{Extra Game One}
In the first stage, Player 1 can choose among three actions: A, B, or C. In the second stage, Player 2 knows when action A was taken, but otherwise cannot tell whether B or C was taken. At this point, Player 2 has three options: D, E, and F. Selecting F ends the game. If Player 2 chooses either D or E, the game advances to the third stage, where Player 1 has a choice between actions G and H. 
Selecting H ends the game and player 1 gets payoff 2 and player 2 gets payoff -1. If Player 1 instead chooses G, Player 2 then decides between actions Q and W, with both players getting 3 after Q, and both players getting 0 after W.

\subsection{Extra Game Two}
In the first stage, Player 1 has three options: A, B, and C. In the second stage, Player 2, without knowing Player 1's choice, can select either D or E. In the third stage, Player 1 has two actions to choose from: F or G. Finally, in the fourth stage, Player 3 can decide between actions Q and W. 
If Player 3 selects Q in the final stage, all players receive a payoff of 3. However, if Player 3 opts for W, Player 1 and Player 2 each receive a payoff of 2, while Player 3 gets 3.

\subsection{Extra Game Three}
A chance node will select one of four possible outcomes: A, B, C, or D. After observing the result of the chance node, Player 1 will choose from three available actions: E, F, or G. In the following stage, Player 2, having observed Player 1's choice, will choose between two actions: Q or W. 
Payoffs are as follows: under A, Q gives (1, -1) and W gives (2, -2); under B, Q gives (3, -3) and W gives (-3, 3); under C, Q gives (0, 0) and W gives (-1, 1); and under D, Q gives (4, -4) and W gives (-4, 4). 

\subsection{Market Entry Model~\cite{levin2002extensive}}
Firm 1 is a monopolist already established in the market. Firm 2, a potential competitor, can choose to enter the market or stay out (if it stays out, the game ends). If Firm 2 enters, Firm 1 must decide whether to compete aggressively (Fight) or allow some market share to Firm 2 (Accommodate). This game only lasts for one round.
The payoffs are structured as follows:
Firm 1 earns more if Firm 2 decides not to enter.
If Firm 2 enters and Firm 1 accommodates, both firms earn an equal amount.
If Firm 2 enters and Firm 1 fights, Firm 2 earns more than Firm 1.

\subsection{Market Signalling Game~\cite{spencer1974market}}
A new manufacturer plans to enter the market, and its strength, determined by chance (not as a player decision), can be either strong (S) with probability 2/3 or weak (W) with probability 1/3. The new manufacturer will then send a signal, either strong (S) or weak (W). The current manufacturer does not know the new manufacturer’s actual strength but observes the signal, which could indicate either strong (S) or weak (W). Based on this signal, the current manufacturer must decide to either compete aggressively (F) or accommodate the new competitor (A). The payoffs for each scenario are as follows:\\
If the new manufacturer is strong and sends a strong signal, and the existing manufacturer chooses to fight, the payoffs are (1, 0) for the new and existing manufacturers, respectively. If the existing manufacturer adapts, the payoffs are (3, 1).\\
If the new manufacturer is strong and sends a weak signal, and the existing manufacturer chooses to fight, the payoff is (0, 0). If the existing manufacturer adapts, the payoff is (2, 1).\\
If the new manufacturer is weak and sends a strong signal, and the existing manufacturer chooses to fight, the payoff is (0, 2). If the existing manufacturer adapts, the payoff is (2, 1).\\
If the new manufacturer is weak and sends a weak signal, and the existing manufacturer chooses to fight, the payoff is (1, 2). If the existing manufacturer adapts, the payoff is (3, 1).

\subsection{Nim (with five in one pile)~\cite{savani2024gambit}}
\label{app:nim_5}
The game involves a single pile of five stones, with two players, Alice and Bob, taking turns. On each turn, a player can remove either one or two stones (but only one stone if only one remains). The goal is to avoid taking the last stone \footnote{The winning condition is adjusted in this misere variation.}.
Here’s how the moves unfold:\\
Alice starts and can choose to remove one or two stones.\\
   If Alice removes one stone, Bob can also remove one or two stones.\\
      \hspace*{0.4cm}If Bob removes one stone, Alice can again choose to remove \hspace*{0.4cm}one or two stones.\\
         \hspace*{0.8cm}If Alice removes one stone, Bob has the choice to remove \hspace*{0.8cm}one or two stones.\\
            \hspace*{1.2cm}If Bob removes one stone, Alice can remove one stone, \hspace*{1.2cm}which would result in Alice losing the game.\\
            \hspace*{1.2cm}If Bob removes two stones, Bob loses the game.\\
         \hspace*{0.8cm}If Alice removes two stones, Bob can remove one stone, \hspace*{0.8cm}leading to Bob's loss.\\
      \hspace*{0.4cm}If Bob removes two stones, Alice can remove one or two stones.\\
         \hspace*{0.8cm}If Alice removes one stone, Bob can remove one stone, \hspace*{0.8cm}leading to Bob's loss.\\
         \hspace*{0.8cm}If Alice removes two stones, Alice loses the game.\\
   If Alice removes two stones, Bob has the option to remove one or two stones.\\
      \hspace*{0.4cm}If Bob removes one stone, Alice can remove one or two stones.\\
         \hspace*{0.8cm}If Alice removes one stone, Bob can remove one stone, \hspace*{0.8cm}leading to Bob's loss.\\
         \hspace*{0.8cm}If Alice removes two stones, Alice loses the game.\\
      \hspace*{0.4cm}If Bob removes two stones, Alice has only one stone left and \hspace*{0.4cm}loses.\\
In this game, the winner earns one point, while the loser loses one point.


\subsection{Nuclear Crisis~\cite{Manea2016}}

After Country B provokes Country A, a sequence of decisions unfolds. Country A must first decide to either escalate (E) the situation or ignore it (I) and end the game. If Country A ignores the provocation, both countries receive a small positive payoff.
If Country A escalates (E), Country B then faces the choice to either back down (B) with a small payoff loss while Country A gains a small payoff or to escalate further (E). 
If both Country A and Country B choose to escalate (E), this brings about a nuclear crisis. Country A then chooses first between retreating (R) and losing credibility or detonating (D). Country B, unaware of Country A’s decision, then decides to retreat (R) and lose credibility or detonate (D).
If one country detonates while the other retreats, the detonating country gains a large positive payoff, while the retreating country incurs a large negative payoff. However, if neither country retreats, retaliation follows the first nuclear strike, resulting in a catastrophic nuclear disaster, yielding a very large negative payoff for both countries. If both countries retreat, they each suffer a small negative payoff.

\subsection{Rock, Paper, Scissors~\cite{sinervo1996rock}}

Both players simultaneously reveal one of three symbols: rock, paper, or scissors. Rock defeats scissors by blunting it, scissors defeat paper by cutting it, and paper defeats rock by covering it.

\subsection{Simple Bargaining Game~\cite{nash1950bargaining}}

A and B are negotiating how to split 10,000 pounds in cash, following these rules: First, A proposes a plan where A receives 5,500 pounds, and B receives 4,500 pounds. If B accepts, the negotiation concludes, with A getting 5,500 pounds and B getting 4,500 pounds. If B rejects, B then proposes an equal split of 5,000 pounds each. If A accepts B’s offer, both players finalize the negotiation with each receiving 4,750 pounds. If A rejects B’s offer, A proposes a new split where A receives 5,200 pounds, and B receives 4,800 pounds. Here, B has no choice but to accept, resulting in A receiving 4,693 pounds and B receiving 4,332 pounds.
The final amount each player receives is their payoff in the negotiation.

\subsection{Tic-Tac-Toe~\cite{crowley1993flexible}}
\label{app:tictactoe}
You’re playing a two-player paper-and-pencil game, where players take turns marking a three-by-three grid with either an ``x'' or an ``o.'' The objective is to get three of your marks in a row, column, or diagonal to win. The winner gains 1 point, while the loser loses 1 point.
Assuming ``e'' represents empty, the current board state is:\\
e | o | e\\
e | x | o\\
x | x | o\\
The next move belongs to ``x.'' 
Here’s how the game tree can develop: \\
"x" can place the mark on (0, 0), (0, 2) and (1, 0).\\
If "x" places the mark on (0, 0), \\
   \hspace*{0.4cm} "o" can then place the mark on (0, 2) and (1, 0).\\
   \hspace*{0.4cm} If "o" places the mark on (0, 2), "o" wins.\\
   \hspace*{0.4cm} If "o" places the mark on (1, 0), \\
      \hspace*{0.8cm} "x" can then place the mark on (0, 2), \\
      \hspace*{0.8cm} If "x" places the mark on (0, 2), "x" wins.\\
If "x" places the mark on (0, 2), "x" wins.\\
If "x" places the mark on (1, 0), \\
   \hspace*{0.4cm} "o" can then place the mark on (0, 0) and (0, 2).\\
   \hspace*{0.4cm} If "o" places the mark on (0, 0), \\
      \hspace*{0.8cm} "x" can then place the mark on (0, 2), \\
      \hspace*{0.8cm} If "x" places the mark on (0, 2), "x" wins.\\
   \hspace*{0.4cm} If "o" places the mark on (0, 2), "o" wins. \\

\lstset{
 breaklines=true, 
 columns=fixed,       
 numbers=left,                                        
 numberstyle=\tiny\color{gray},                       
 frame=none,                                          
 backgroundcolor=\color[RGB]{245,245,244},            
 keywordstyle=\color[RGB]{40,40,255},                 
 numberstyle=\footnotesize\color{darkgray},           
 commentstyle=\it\color[RGB]{0,96,96},                
 stringstyle=\rmfamily\slshape\color[RGB]{128,0,0},   
 showstringspaces=false,                              
 language=python,                                        
}
\onecolumn
\section{Demonstrations in Prompts}
In this section, we present the demonstrations incorporated into the prompt template shown in Table~\ref{tab:prompt}.
\subsection{Code Generation Initialization}
For the code generation initialization prompts, we include two code examples along with the API documentation. The complete pygambit API documentation, which includes the examples we reference, is accessible at: \url{https://gambitproject.readthedocs.io/en/latest/pygambit.api.html}.

\noindent
\codeexampleone

\noindent
\textbf{Game description:}

\noindent
There are two players, a Buyer and a Seller. The Buyer moves first and has two actions, Trust or Not trust. If the Buyer chooses Not trust, then the game ends, and both players receive payoffs of 0. If the Buyer chooses Trust, then the Seller has a choice with two actions, Honor or Abuse. If the Seller chooses Honor, both players receive payoffs of 1; if the Seller chooses Abuse, the Buyer receives a payoff of -1 and the Seller receives a payoff of 2.

\noindent
\textbf{Code:}
\begin{lstlisting}[numbers=none]
import pygambit as gbt
g = gbt.Game.new_tree(players=["Buyer", "Seller"], title="One-shot trust game")
g.append_move(g.root, "Buyer", ["Trust", "Not trust"])
g.append_move(g.root.children[0], "Seller", ["Honor", "Abuse"])
g.set_outcome(g.root.children[0].children[0], g.add_outcome([1, 1], label="Trustworthy"))
g.set_outcome(g.root.children[0].children[1], g.add_outcome([-1, 2], label="Untrustworthy"))
g.set_outcome(g.root.children[1], g.add\_outcome([0, 0], label="Opt-out"))
#Save the EFG
efg = g.write(format='native')
\end{lstlisting}

\noindent
\codeexampletwo

\noindent
\textbf{Game description:}

\noindent
There are two players, Alice and Bob. There is a deck of cards, with equal numbers of King and Queen cards. The game begins with each player putting \$1 in the pot. One card is dealt at random to Alice; Alice observes her card but Bob does not. After Alice observes her card, she can choose either to Raise or to Fold. If she chooses to Fold, Bob wins the pot and the game ends. If she chooses to Raise, she adds another \$1 to the pot. Bob then chooses either to Meet or Pass. If he chooses to Pass, Alice wins the pot and the game ends. If he chooses to Meet, he adds another \$1 to the pot. There is then a showdown, in which Alice reveals her card. If she has a King, then she wins the pot; if she has a Queen, then Bob wins the pot.

\noindent
\textbf{Code:}
\begin{lstlisting}[numbers=none]
import pygambit as gbt
g = gbt.Game.new_tree(players=["Alice", "Bob"], title="One card poker game")
g.append_move(g.root, g.players.chance, ["King", "Queen"])
for node in g.root.children:
    g.append_move(node, "Alice", ["Raise", "Fold"])
g.append_move(g.root.children[0].children[0], "Bob", ["Meet", "Pass"])
g.append_move(g.root.children[1].children[0], "Bob", ["Meet", "Pass"])
# Set infoset
g.set_infoset(g.root.children[0].children[0], g.root.children[1].children[0].infoset)
alice_winsbig = g.add_outcome([2, -2], label="Alice wins big")
alice_wins = g.add_outcome([1, -1], label="Alice wins")
bob_winsbig = g.add_outcome([-2, 2], label="Bob wins big")
bob_wins = g.add_outcome([-1, 1], label="Bob wins")
g.set_outcome(g.root.children[0].children[0].children[0], alice_winsbig)
g.set_outcome(g.root.children[0].children[0].children[1], alice_wins)
g.set_outcome(g.root.children[0].children[1], bob_wins)
g.set_outcome(g.root.children[1].children[0].children[0], bob_winsbig)
g.set_outcome(g.root.children[1].children[0].children[1], alice_wins)
g.set_outcome(g.root.children[1].children[1], bob_wins)
# Save the EFG
efg = g.write(format='native')
\end{lstlisting}

\noindent
\api
\begin{enumerate}

\item  pygambit.gambit.Game.append\_move

\begin{lstlisting}[numbers=none]
Game.append\_move(nodes: Node | str | Iterable[Node | str], player: Player | str, actions: List[str]) -> None
\end{lstlisting}

Add a move for player at terminal nodes. All elements of nodes become part of a new information set, with actions labeled according to actions.

\item pygambit.gambit.Game.add\_outcome
\begin{lstlisting}[numbers=none]
Game.add_outcome(payoffs: List | None = None, label: str = '') -> Outcome
\end{lstlisting}
Add a new outcome to the game.

\item  pygambit.gambit.Game.set\_outcome
\begin{lstlisting}[numbers=none]
Game.set_outcome(node: Node | str, outcome: Outcome | str | None) -> None
\end{lstlisting}
Set outcome to be the outcome at node. If outcome is None, the outcome at node is unset.

\item pygambit.gambit.Game.set\_infoset
\begin{lstlisting}[numbers=none]
Game.set_infoset(node: Node | str, infoset: Infoset | str) -> None
\end{lstlisting}
Place node in the information set infoset. node must have the same number of descendants as infoset has actions.
You can set a node's infoset using the following code:
\begin{lstlisting}[numbers=none]
g.set_infoset(node1, node2.infoset)
\end{lstlisting}
Note that when grouping more than two nodes into the same information set, you should assign all nodes to the infoset of one designated node. For example, if you have three nodes—node1, node2, and node3—in the same information set, you can set it up as follows:
\begin{lstlisting}[numbers=none]
g.set_infoset(node1, node2.infoset)
g.set_infoset(node3, node2.infoset)
\end{lstlisting}

\item pygambit.gambit.Game.set\_chance\_probs
\begin{lstlisting}[numbers=none]
Game.set_chance_probs(infoset: Infoset | str, probs: Sequence)
\end{lstlisting}
For example: 
\begin{lstlisting}[numbers=none]
g.append_move(g.root, g.players.chance, [f"Chance {i}" for i in range(num_choices)])
probabilities = np.random.randint(1,20,n)
total = sum(probabilities)
# Set all the probabilities for the chance node at the same time.
g.set_chance_probs(g.root.infoset, [gbt.Rational(p,total) for p in probabilities])
\end{lstlisting}

\end{enumerate}

\subsection{Imperfect Information Retrieval Initialization}

For the initialization of imperfect information retrieval, we utilize three demonstrations, each comprising a game description, a thought process, and the resulting output.

\noindent
\imone

\noindent
\textbf{Game description:}

\noindent
There are two players in this game. 
Player 1 and Player 2 move simultaneously. Player 1 has two possible actions: A and B. Player 2 also has two possible actions: X and Y.

\noindent
\textbf{Reasoning process:}

\noindent
Since the players move simultaneously, this game involves imperfect information. 
To model the game with a sequential structure, assume Player 2 moves first and Player 1 moves second.
The starting node, g.root, is Player 2's decision node.
Player 2 has two actions, X and Y. So, Player 1's decision nodes are g.root.children[0] and g.root.children[1], which correspond to Player 2's actions X and Y.
Player 1, moving simultaneously, does not know which action Player 2 has taken. 
Therefore, Player 1's decision nodes should be grouped in the same information set.

\noindent
\textbf{Output:}
\begin{lstlisting}[numbers=none]
g.set_infoset(g.root.children[0], g.root.children[1].infoset)
\end{lstlisting}

\noindent
\imtwo

\noindent
\textbf{Game description:}

\noindent
There are two players, Alice and Bob. A deck of cards contains an equal number of Kings and Queens. The game begins with each player contributing \$1 to the pot. One card is randomly dealt to Alice, who observes it while Bob does not. After observing her card, Alice can choose either to Raise or Fold. If she folds, Bob wins the pot. If Alice raises, she adds another \$1 to the pot, and Bob then decides either to Meet or Pass. If Bob passes, Alice wins the pot; if Bob meets, a showdown occurs. If Alice holds a King, she wins the pot. If she holds a Queen, Bob wins the pot.

\noindent
\textbf{Reasoning process:}

\noindent
This game involves imperfect information because Bob does not know Alice's card.
The game begins with a chance node (g.root) that randomly assigns a King or Queen to Alice and Bob.
Alice moves first. Since the chance node has two outcomes (King and Queen), Alice's decision nodes are g.root.children[0] (King) and g.root.children[1] (Queen). Alice can Raise or Fold.
If Alice raises, Bob must decide whether to Meet or Pass. Bob's decision nodes are g.root.children[0].children[0] (King and Raise) and g.root.children[1\\].children[0] (Queen and Raise). 
Since Bob cannot distinguish between Alice holding a King or a Queen, Bob's decision nodes should be grouped in the same information set.

\noindent
\textbf{Output:}
\begin{lstlisting}[numbers=none]
g.set_infoset(g.root.children[0].children[0], g.root.children[1].children[0].infoset)
\end{lstlisting}

\noindent
\imthree

\noindent
\textbf{Game description:}

\noindent
This is a two players card game between Alice and Bob. Nature has 3 cards, Jack, Queen, and King. A chance node will draw one card for Alice, one card for Bob. Alice knows her own card but not Bob's card. If she bets and her card is higher than Bob's card she wins 1, if lower -1. If she does not bet, she gets 0.

\noindent
\textbf{Reasoning process:}

\noindent
There are total 6 possible outcomes for the chance node: 
JQ (g.root.children[0]), JK (g.root.children[1]), QJ (g.root.children[2]), QK (g.root.children[3]), KJ (g.root.children[4]), KQ (g.root.children[5]).\\
Alice have the same card situations are (grouping the card situations that are the same for Alice): 
JQ (g.root.children[0]) and JK (g.root.children[1]), QJ (g.root.children[2]) and QK (g.root.children[3]), KJ (g.root.children[4]) and KQ (g.root.children[5]).\\
Bob have the same card situations are (grouping the card situations that are the same for Bob):
QJ (g.root.children[2]) and KJ (g.root.children[4]), JQ (g.root.children[0]) and KQ (g.root.children[5]), JK (g.root.children[1]) and QK (g.root.children[3]).\\
These nodes are the decision nodes for Alice, because she only knows her own card, we have to group JQ and JK, QJ and QK, KJ and KQ in the same information set.

\noindent
\textbf{Output:}
\begin{lstlisting}[numbers=none]
g.set_infoset(g.root.children[0], g.root.children[1].infoset)
g.set_infoset(g.root.children[2], g.root.children[3].infoset)
g.set_infoset(g.root.children[4], g.root.children[5].infoset)
\end{lstlisting}
\subsection{Imperfect Information Retrieval}
The game description in this section serves as the test case for translating natural language into an EFG representation. It can be one of the descriptions from Appendix~\ref{app:game}  or any other game descriptions found in textbooks or online sources.
\subsection{EFG Generation}\label{app: stage 1 output}
The game description is identical to the one in the previous section. Additionally, there is code for imperfect information, which corresponds to the output from stage one in Figure~\ref{fig:pipeline}. Furthermore,  this section of the prompts also outlines specific constraints for the generated code.

\noindent
In the case of an imperfect information game, the output of imperfect retrieval will include both the reasoning in the form of code comments and the code required to set up the information set. Below is a sample output for stage one of a Bagwell game, generated by GPT-4o.

\noindent
\imperfect

\begin{lstlisting}[numbers=none]
"""
Step-by-step Reasoning process:

1. The game starts with Player 1 making a decision between "S" and "C".
2. Player 2 then attempts to discern Player 1's choice, but there is a small chance of misinterpretation.
3. This misinterpretation introduces imperfect information for Player 2.
4. We need to model this by creating a chance node that represents the misinterpretation.
5. Player 2's decision nodes will be based on what they believe Player 1 chose.
6. Since Player 2 cannot perfectly discern Player 1's choice, their decision nodes should be grouped in the same information set.

Assuming the chance node has two outcomes: Player 1's choice is correctly interpreted or misinterpreted.
Player 2's decision nodes are based on these interpretations.

Group Player 2's decision nodes in the same information set to reflect the imperfect information.
"""
g.set_infoset(g.root.children[0].children[0], g.root.children[1].children[0].infoset)
g.set_infoset(g.root.children[0].children[1], g.root.children[1].children[1].infoset)
\end{lstlisting}

\noindent
For perfect information games, the output generated will include only code comments. These comments will highlight that setting information sets for the given game is unnecessary. An example output generated by GPT-4o for the Market Entry Model Game is provided below.

\noindent
\imperfect

\begin{lstlisting}[numbers=none]
```python
# In this game, Firm 2 decides first whether to enter or not.
# If Firm 2 enters, Firm 1 then decides how to compete.
# Firm 1's decision nodes are g.root.children[0] (Enter) and g.root.children[1] (Out).
# Since Firm 1 knows whether Firm 2 has entered or not, there is no imperfect information for Firm 1.
# Therefore, there is no need to set any information sets in this game.
```
\end{lstlisting}

\noindent
\constraints

\noindent
Avoid using recursion or loops in the code generation.  

\noindent
Refrain from using the ``+'' operator in the code generation.  

\noindent
Note that `g.append\_move', `g.append\_infoset', `g.set\_outcome', and `g.set\_infoset' all return `None'.  

\noindent
Make sure to call `g.set\_infoset(node1, node2.infoset)' only after `g.append\_move' has been executed on both `node1' and `node2'.  

\noindent
Use `gbt.Rational' to represent probabilities at the chance node. 

\subsection{Error Message}
The prompts used in the self-debugging stage consist of two components: the system's error message and instructions for resolving common bugs. Below is an example of a potential bug in pygambit along with the corresponding error message from the interpreter.

\noindent
\error

\noindent
name `node' is not defined

\noindent
\derror

\noindent
Functions like g.append\_move, g.append\_infoset, g.set\_outcome, and g.set\_infoset return None.

\noindent
Regarding the error ``name `node' is not defined'': Regenerate the path for node1 and node2 when using g.set\_infoset(node1, node2.infoset).


\end{document}